\documentclass[11pt]{article}

\usepackage[final]{acl}
\usepackage{xcolor}
\usepackage[table]{xcolor}
\definecolor{blackcolor}{rgb}{0.95,0.95,0.95}

\usepackage{times}
\usepackage{pgfplots}
\pgfplotsset{compat=1.18}
\usepackage{latexsym}
\usepackage{booktabs} 
\usepackage{makecell} 
\usepackage{graphicx}
\usepackage{multirow}
\usepackage{pifont}
\usepackage{tcolorbox}
\tcbuselibrary{skins}
\newcommand{\cmark}{\checkmark}
\newcommand{\xmark}{\ding{55}}
\usepackage{amssymb}
\usepackage{microtype}
\usepackage{sidecap}

\newcommand{\findingz}[2]{
    \vspace{0.5em}
    \begin{tcolorbox}[
        colback=blue!5,
        colframe=blue!60!black,
        arc=2pt,
        boxsep=2pt,
        left=5pt, right=5pt,
        top=2pt, bottom=2pt,
        boxrule=0.6pt,
        drop shadow={opacity=0.25},
        enhanced jigsaw
    ]
    \textbf{\textit{Finding #1:}} #2
    \end{tcolorbox}
    \vspace{0.3em}
}

\usepackage{listings}
\usepackage[T1]{fontenc}

\usepackage[utf8]{inputenc}

\usepackage{pgfplots}
\usepackage{pgf-pie}
\pgfplotsset{compat=1.18}

\usepackage{subcaption}

\usepackage{microtype}
\usepackage{soul}
\usepackage{enumitem}
\usepackage{inconsolata}

\usepackage{graphicx}

\usepackage{wrapfig}
\title{\raisebox{-0.2\height}{\includegraphics[height=1.3em]{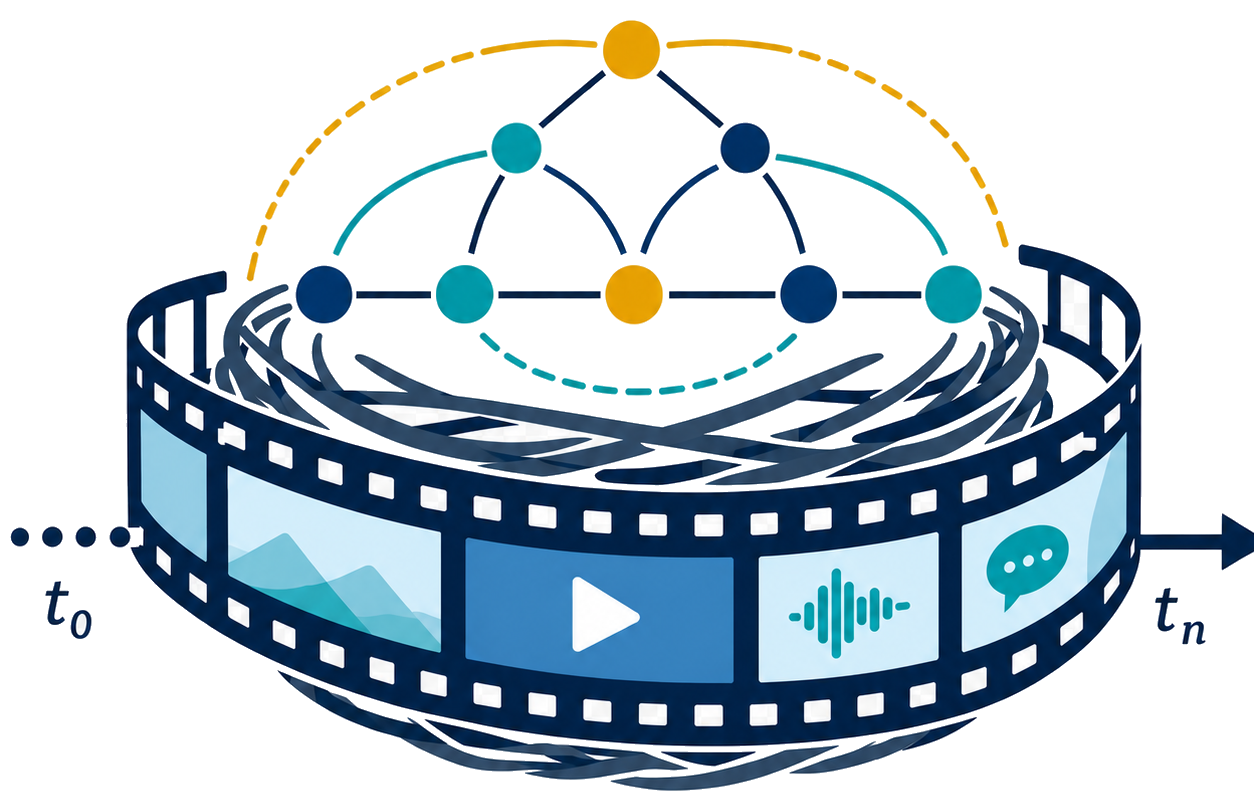}}\,NEST: Narrative Event Structures in Time for Long Video Understanding}

\author{
  \textbf{Ali Asgarov} \quad
  \textbf{Kaushik Narasimhan} \quad
  \textbf{Najibul Haque Sarker} \quad
  \textbf{Hani Alomari} \\
  \textbf{Chia-Wei Tang} \quad
  \textbf{Anushka Sivakumar} \quad
  \textbf{Zaber Ibn Abdul Hakim} \\
  \textbf{Shaurya Mallampati} \quad
  \textbf{Chris Thomas} \\[4pt]
  Department of Computer Science, Virginia Tech \\
  Blacksburg, VA, USA \\
  \texttt{aliasgarov@vt.edu}
}

\begin{document}
\maketitle
\begin{abstract}
Recent progress in vision-language models has enabled processing of increasingly long video sequences, but handling extended token streams does not translate to understanding complex narrative structure in long videos. Existing long-video benchmarks focus on needle-in-a-haystack retrieval rather than evaluating how low-level actions form events, interact across time, and drive narratives, for example whether a model can connect an early job loss to a later relationship breakup despite intervening scenes or flashbacks. We introduce NEST (Narrative Event Structures in Time for Long Video Understanding), a dataset of 1,005 full-length movies (avg. 98 minutes), each annotated with about 103 multimodal narrative events grounded in visual content, dialogue, and audio. NEST links these events through relations that reflect narrative structure, including explicit temporal ordering, hierarchical composition, and long-range dependencies. We introduce baselines for event trigger detection (ETD), event localization (EL), event argument extraction (EAE), and event relation extraction (ERE). The benchmark is highly challenging for grounded event discovery, with the best performance across evaluated zero-shot models on main evaluation subsets remaining below 8\% for ETD, 6\% for EL, and 11\% for EAE. In contrast, ERE is more tractable once events are given, reaching 34.23\% F1 zero-shot and 49.20\% F1 after fine-tuning.

\end{abstract}
 
\begin{figure*}
    \centering
   \includegraphics[width=1\linewidth]{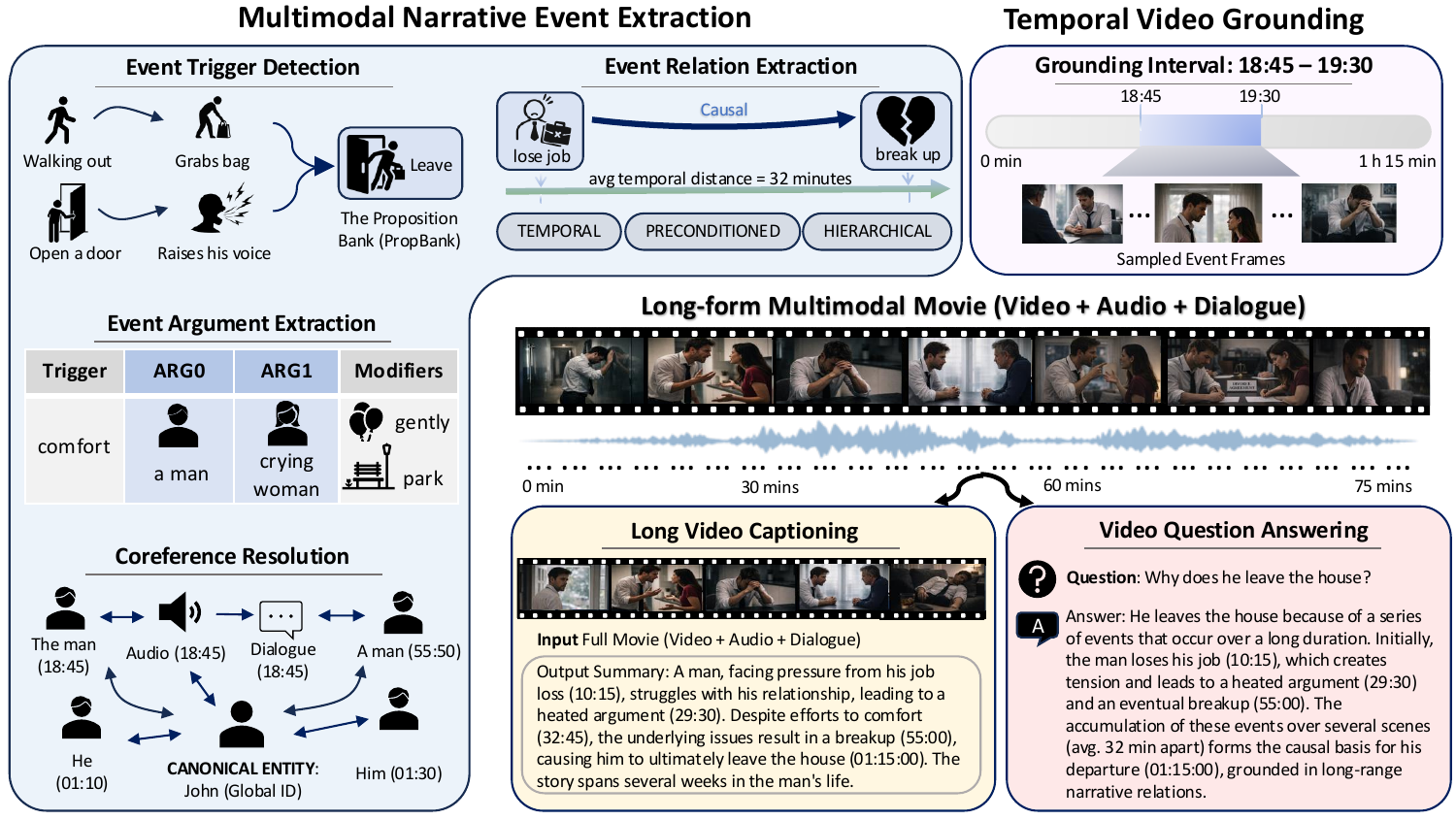}
    \caption{NEST evaluates four narrative event tasks on full-length movies: Event Trigger Detection (ETD), Event Localization (EL), Event Argument Extraction (EAE), and Event Relation Extraction (ERE). Annotations are grounded in visual content, dialogue, and audio, and linked through relations that reflect narrative structure.}
    \label{fig:NEST}
\end{figure*}

\section{Introduction}

\begin{table*}[t]
\centering
\scriptsize
\resizebox{1\textwidth}{!}{%
\begin{tabular}{lcccccccccc}
\toprule
\textbf{Dataset} & \textbf{\# Hrs} & \textbf{Avg.} & \textbf{Ann.} & \textbf{Eval.} & \textbf{Source Availability} & \textbf{Event} & \textbf{ERE} & \textbf{Multi-R} & \textbf{Audio} \\    
\midrule
CinePile \cite{rawal2024cinepilelongvideoquestion} & 418 & 2.67 & Auto/Man. & MC & YT links & \checkmark & \checkmark & \xmark & \cmark \\
EgoSchema \cite{3666122.3668126} & 253.2 & 3.00 & Auto/Man. & MC & Videos & \checkmark & \xmark & \xmark & \xmark \\
EgoPlan-Bench2 \cite{qiu2025egoplanbench2benchmarkmultimodallarge} & 92.8 & $\leq$5 & Auto/Man. & MC & Videos & \xmark & \checkmark & \xmark & \xmark \\
LongVideoBench \cite{wu2024longvideobench} & 494.9 & 7.89 & Manual & MC & Videos & \checkmark & \checkmark & \cmark & \xmark \\
Video-MMMU \cite{hu-etal-2026-video} & 42.2 & 8.44 & Manual & MC & Videos & \xmark & \xmark & \xmark & \checkmark \\
MovieChat-1K \cite{10657734} & 156.7 & 9.40 & Manual & MC+OE & Videos & \checkmark & \xmark & \checkmark & \xmark \\
MLVU \cite{MLVU} & 346 & 12.00 & Auto/Man. & MC+OE & Videos & \checkmark & \checkmark & \checkmark & \xmark \\
Neptune \cite{nagrani2025neptunelongorbitbenchmarking} & 601.3 & $\leq$15 & Auto/Man. & MC+OE & Videos & \checkmark & \checkmark & \checkmark & \checkmark \\
Video-MME (Long) \cite{fu2025videommefirstevercomprehensiveevaluation} & 596.4 & 39.76 & Manual & MC & YT links & \checkmark & \checkmark & \checkmark & \checkmark \\
HourVideo \cite{chandrasegaran2024hourvideo} & 382.5 & 45.70 & Auto/Man. & MC & Videos & \checkmark & \checkmark & \checkmark & \xmark \\
InfiniBench \cite{ataallah-etal-2025-infinibench} & 1066.3 & 52.59 & Auto/Man. & MC+OE & Key frames & \checkmark & \checkmark & \checkmark & \checkmark \\
LVBench \cite{wang2025lvbenchextremelongvideo}& 117.3 & 68.35 & Manual & MC & YT links & \checkmark & \checkmark & \checkmark & \xmark \\
MF$^2$ \cite{zaranis2025mf2} & 78.0 & 88.33 & Manual & Claim pairs & Videos & \checkmark & \checkmark & \checkmark & \checkmark \\
\midrule
\textbf{NEST} & \textbf{1639.3} & \textbf{97.87} & \textbf{Auto/Man.} & \textbf{MC+OE} & \textbf{Videos | Video \& Image \& Audio Features} & \textbf{\checkmark} & \textbf{\checkmark} & \textbf{\checkmark} & \textbf{\checkmark} \\
\bottomrule
\end{tabular}%
}
\caption{Comparison of long-video understanding benchmarks. Columns report duration, annotation type, evaluation format (MC = multiple-choice, OE = open-ended), source availability, event understanding, ERE, multi-scene reasoning, and audio usage.}
\label{tab:longvideo}
\vspace{-2em}
\end{table*}

Video has become the dominant medium for information and storytelling. Its blend of language, vision, and sound makes it deeply engaging, but understanding video narratives requires more than processing individual frames. Recent vision-language models (VLMs) have demonstrated strong performance across tasks involving images and videos \cite{11092508, chen2024internvl, bai2025qwen25vltechnicalreport, zhang2025videollama3frontiermultimodal, xu2025qwen25omnitechnicalreport, zhang2025llavavideo, li2025ariaopenmultimodalnative, 11094148, guo2025seed15vltechnicalreport}, but most struggle with the deeper narrative structure of long-form video. These systems treat video as a flat stream of tokens and lack scaffolding to reason over meaningful abstractions separated by long temporal distances. This limitation becomes particularly important when understanding a narrative requires connecting events that unfold across distant scenes. As models scale, long-form video understanding with narrative-level reasoning becomes essential for education, storytelling, and video analysis.

\textbf{\textit{Current benchmarks remain limited in their ability to assess narrative-level understanding over long videos}.} Video event understanding benchmarks \cite{9577635, khan2022grounded} focus on atomic actions in short clips, which cannot capture narrative significance. Events have complex hierarchical structures \cite{RADVANSKY2017133}. For example, ``\textit{person leaving home}'' decomposes into ``\textit{grabbing keys}'', ``\textit{opening door}'', and ``\textit{walking out}'', yet these atomic actions alone miss the narrative meaning of \textit{why} someone is leaving. While some work addresses narrative-level events \cite{liang2025videventlargedatasetunderstanding}, they operate on short video clips and fail to assess how understanding one event requires knowledge of another from much earlier in the narrative. Existing long-video benchmarks \cite{rawal2024cinepilelongvideoquestion, wu2024longvideobench, hu-etal-2026-video, chandrasegaran2024hourvideo, ataallah-etal-2025-infinibench} emphasize other aspects over narrative event extraction, often falling short in duration, scale, or annotation quality (Table~\ref{tab:longvideo}), and many rely on multiple-choice formats \cite{fu2025videommefirstevercomprehensiveevaluation, wang2025lvbenchextremelongvideo, hu-etal-2026-video} with selection biases \cite{3db92d5885ab4ed4a6fd18bb5edf8636, singh2025optionspitfallsmultiplechoicequestions} or target retrieval \cite{wang-etal-2025-multimodal, zhao2025needle} rather than narrative understanding.

\textbf{\textit{These limitations highlight the need for models that can reason over hours-long video content.}}

Our contributions are:

\begin{itemize}[noitemsep,nolistsep]

\item We introduce NEST\footnote{\url{https://github.com/aliasgerovs/nest}}, a benchmark for narrative understanding in 1,005 full-length movies (avg. 98 min), with subtitles and structured narrative event annotations.

\item We introduce a multi-task framework covering event trigger detection, event argument extraction, temporal localization, and event relation extraction beyond multiple-choice formats.

\item We evaluate state-of-the-art models, showing that narrative understanding remains challenging for long-video models.

\item We release video and audio features, code, and fine-tuned model checkpoints trained on NEST.

\end{itemize}

\section{Related Work}

\paragraph{Long-context vision models and benchmarks.}
Long-context models can now handle increasingly large context windows \cite{chen2023extendingcontextwindowlarge, peng2024yarn, grattafiori2024llama3herdmodels, bai-etal-2024-longalign, abdin2024phi-}, enabling visual reasoning over extended sequences \cite{3600270.3601993, NEURIPS2023_6dcf277e, bai2023qwenvlversatilevisionlanguagemodel, wang2022internvideogeneralvideofoundation, 10657734, wang2024videotree, chen2025grounded}. While these systems excel at retrieval and search over long contexts \cite{hsieh2024ruler, bai-etal-2024-longalign, NEURIPS2025_8190b210}, they fail at reasoning over complex event relationships \cite{fang-etal-2024-complex, zhang2024structuredeventreasoninglarge}. Benchmarks have driven progress in temporal reasoning over short clips \cite{nextqa, wu2021star} and domain-specific settings \cite{3666122.3668126, qiu2025egoplanbench2benchmarkmultimodallarge}, but most focus on content under three minutes. Longer benchmarks \cite{wu2024longvideobench, chandrasegaran2024hourvideo, ataallah-etal-2025-infinibench} fall short in duration or quality, and most use biased multiple-choice formats \cite{fu2025videommefirstevercomprehensiveevaluation, wang2025lvbenchextremelongvideo}. Neptune \cite{nagrani2025neptunelongorbitbenchmarking} uses free-form answers but stays limited to 15 minutes. \textbf{\textit{These benchmarks emphasize retrieval over reasoning about event structures and causal dependencies.}}

\paragraph{Event understanding datasets.}

Event Extraction (EE) identifies event types, triggers, and arguments through end-to-end \cite{luan-etal-2019-general, lin-etal-2020-joint, huang-peng-2021-document} or pipeline methods \cite{liu-etal-2020-event, du-cardie-2020-event}. Multimodal EE extracts from images and videos~\cite{chen-etal-2021-joint-multimedia-event, sanders-etal-2024-grounding, li-etal-2020-cross, xing-etal-2025-benchmarking}. M2E2~\cite{li-etal-2020-cross} jointly extracts events and arguments from text and images, and Video M2E2~\cite{chen-etal-2021-joint-multimedia-event} extends this to short news videos with multimodal event coreference. \citet{sanders-etal-2024-grounding} further model hierarchical and coreference structure across multiple videos, but NEST differs by modeling long-range event structure within a single continuous full-length narrative. MovieGraphs~\cite{vicol2018moviegraphsunderstandinghumancentricsituations} annotates human-centric situations in movie clips with interaction graphs but operates on short scenes rather than full-length narrative arcs. ImSitu \cite{7780966} extracts structured representations from images, while VidSitu \cite{9577635} and Grounded VidSitu \cite{khan2022grounded} extract from short clips. These works build on PropBank \cite{palmer-etal-2005-proposition}, a semantic role labeling resource that links verbs to predicate senses and structured argument roles (e.g., ARG0 for agent, ARG1 for patient, ARGM-LOC for location), which we also adopt to constrain our event ontology. \textbf{\textit{Most prior video event benchmarks focus on atomic events in short, seconds-long clips rather than event structures that unfold across a continuous full-length narrative.}} VidEvent \cite{liang2025videventlargedatasetunderstanding} extends from seconds to minutes addressing narrative-level events, but operates on short clips, while NEST substantially extends the temporal horizon and event scale (Table~\ref{tab:dataset-stats}). Prior work does not address the same setting of tracing how events compose into broader narrative arcs and causal chains within a single continuous full-length movie. NEST addresses this gap by requiring models to extract and relate high-level events across full-length movies, assessing understanding of semantic hierarchies and causal relationships spanning hours. More closely related benchmarks differ from NEST either in temporal horizon or task formulation. VidEvent~\cite{liang2025videventlargedatasetunderstanding} uses a similar structured event schema but operates on short movie recaps averaging approximately 105s, whereas NEST applies structured trigger, argument, and relation grounding to full-length movies averaging approximately 98 minutes. MF$^2$~\cite{zaranis2025mf2} also contains full-length films but formulates understanding as true/false claim verification rather than structured event extraction. MovieRecapsQA~\cite{shaar2026movierecapsqamultimodalopenendedvideo} evaluates open-ended question answering over condensed movie recaps. Thus, NEST's contribution is not a new event-extraction formulation, but extending structured event grounding and relation reasoning to continuous full-length narratives with long-range dependencies.

\begin{figure}
    \centering
    \includegraphics[width=1\linewidth]{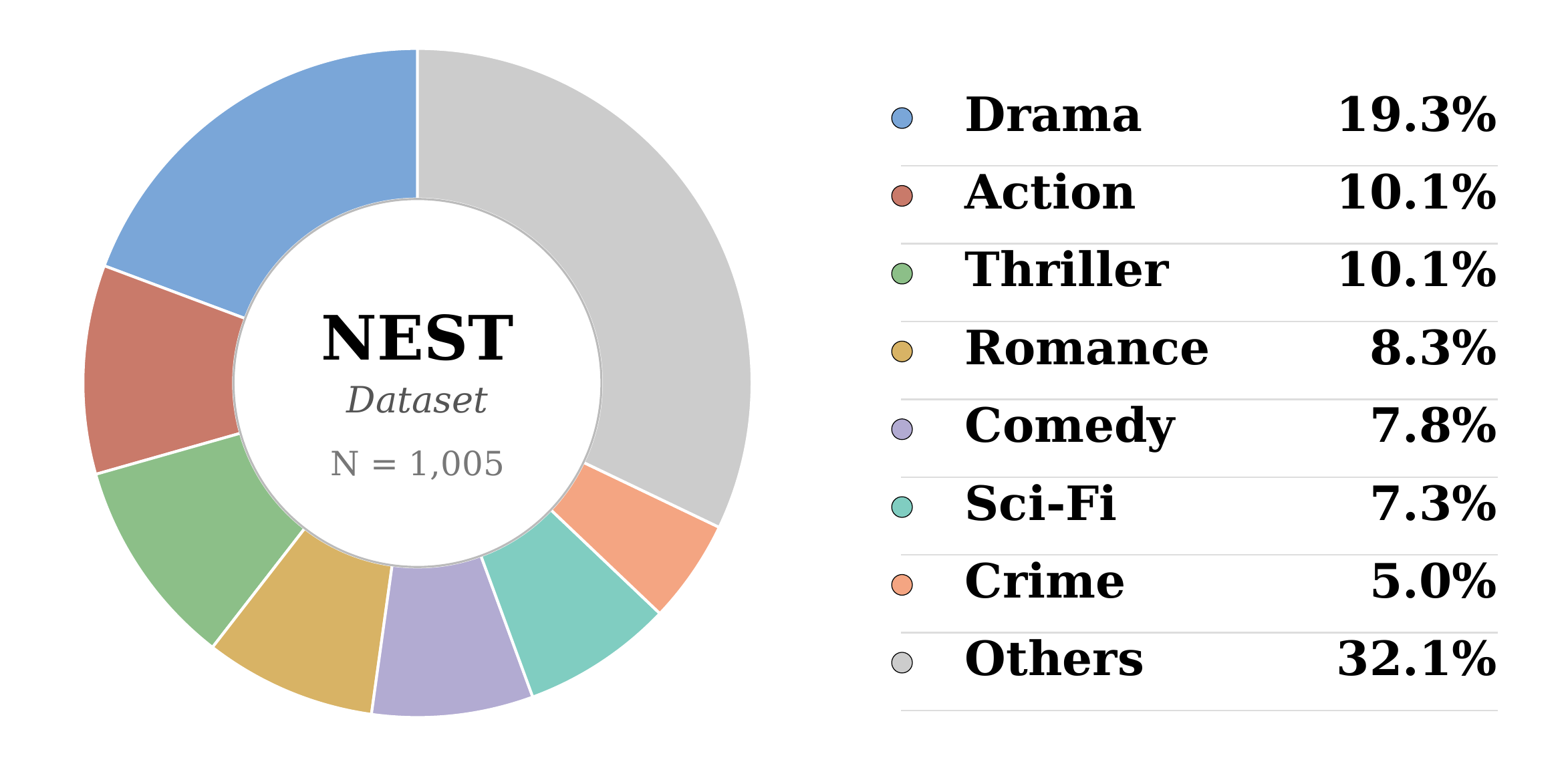}
    \caption{Genre distribution across the 1,005 movies in the NEST dataset.}
    \label{fig:Genre distribution}
\end{figure}

\begin{table}
\centering
\small
\begin{tabular}{lc}
\toprule
\textbf{Statistic} & \textbf{Value} \\
\midrule
Avg. Movie Duration & 97.87 min \\
Avg. Plot Length & 554.09 words \\
Avg. \# Scenes per Movie & 160.61 \\
Avg. Words per Scene Audio Description & 85.52 \\
\bottomrule
\end{tabular}
\caption{Scene-level, duration, and text-length statistics for movies in the NEST dataset.}
\label{tab:scene-duration}
\end{table}

\begin{table*}[t]
\centering
\resizebox{\textwidth}{!}{%
\begin{tabular}{lccccccc}
\toprule
\textbf{Dataset} & \textbf{\# Videos} & \textbf{Avg. Video Len. (s)} & \textbf{Total Hours} & \textbf{Events / Video} & \textbf{Args / Event} & \textbf{Relations / Movie} & \textbf{Avg. Temp. Dist. between Events} \\
\midrule
VidSitu \cite{9577635} \& GVSR \cite{khan2022grounded}
& 29,200 & 10.0 & 81.1 & 6.58 & 3.83 & 3.62 & 3.2 sec \\

VidEvent \cite{liang2025videventlargedatasetunderstanding}
& 1,068 & 105.4 & 31.3 & 21.35 & 3.48 & 16.11 & 36.30 sec \\
\midrule
\textbf{NEST (Ours)}
& 1,005 & \textbf{5,872.2} & \textbf{1,639.3} & \textbf{102.69} & 2.95 & \textbf{5100.00} & \textbf{1920 sec} \\
\bottomrule
\end{tabular}%
}
\caption{Comparison of dataset statistics for video event extraction benchmarks. Per-video, per-event, and temporal-distance statistics are arithmetic means. GVSR \cite{khan2022grounded} shares the same underlying dataset as VidSitu \cite{9577635}.}
\label{tab:dataset-stats}
\end{table*}

\begin{figure*}
    \centering
    \includegraphics[width=1\textwidth]{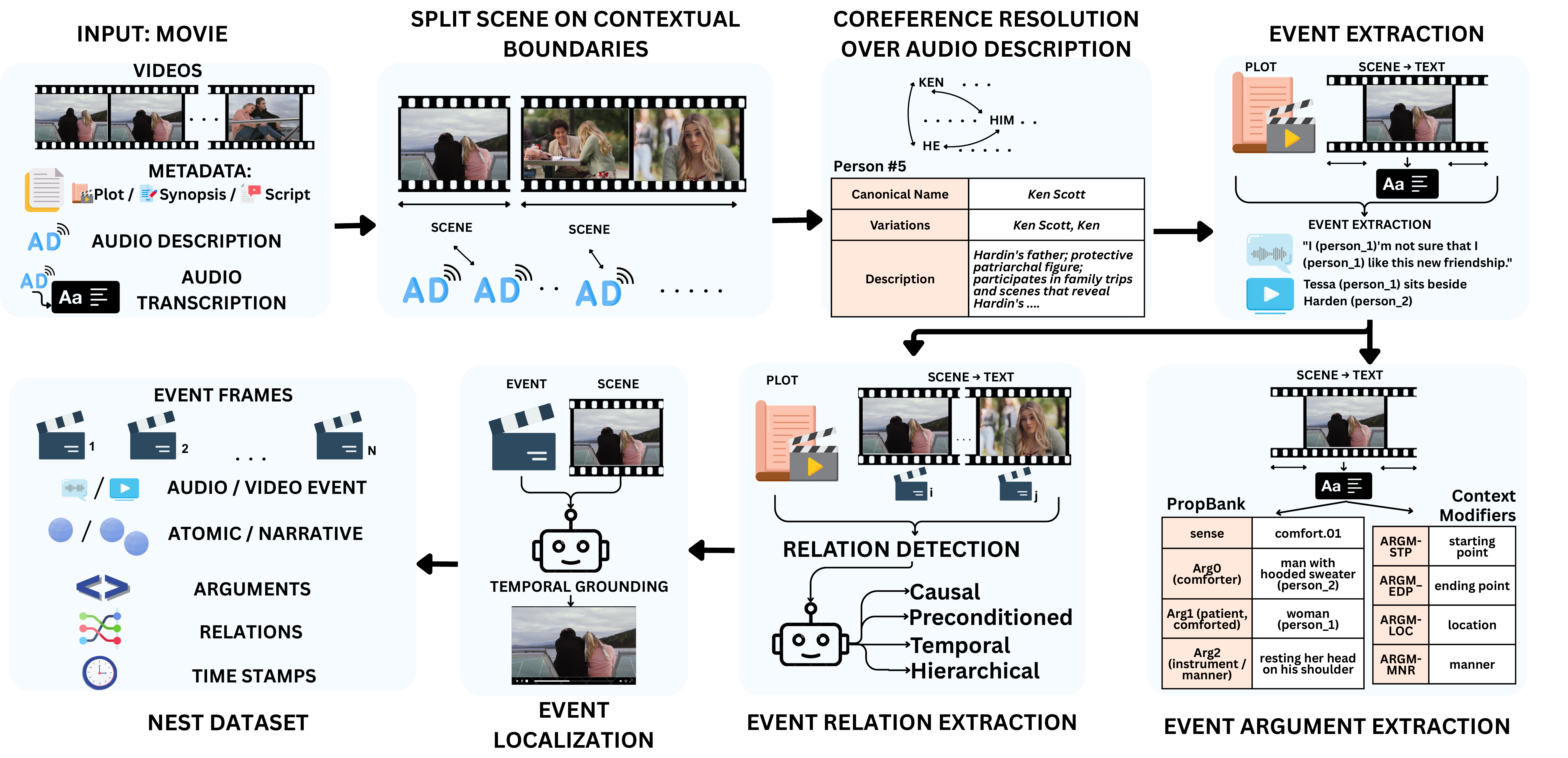}
    \caption{Full-length movies with plots/scripts and audio descriptions (AD) are segmented into scenes at contextual boundaries, and we run coreference on AD to build a unified entity table. For each scene, we extract audio and visual events, recover PropBank arguments and modifiers, and link events with relations. A temporal grounding stage localizes every event to timestamps, giving specific event frames that form the NEST dataset.}

    \label{fig:Full-length movies}
    \vspace{-1em}
\end{figure*}

\section{NEST}

NEST builds upon existing movie datasets \cite{10.1007/978-3-030-58548-8_41, 7780870} and harvests additional open-domain videos at least one hour long from the Library of Congress \cite{loc_collections_films_videos}, archive.org, PublicDomainMovies \cite{publicdomainmovie}, YouTube, and other databases under fair use for research \cite{fairuse}, resources not previously used for long-form video understanding. NEST consists of an automatically labeled training set and a large automatically labeled evaluation set, together with a densely human-annotated GOLD subset.

\subsection{Data Collection and Processing}
We collected metadata including plot summaries, synopses, and scripts from IMDb, Wikipedia, OpenSubtitles, and existing datasets. Audio Description (AD) tracks from AudioVault \cite{audiovault} provided professional narration of visual content. We transcribed these using Whisper \cite{3618408.3619590} and employed LLMs to fix errors and improve alignment. For coreference resolution, we assigned unique identifiers to each entity to maintain consistency throughout videos, using Maverick \cite{martinelli-etal-2024-maverick} and LLM-based methods \cite{gan-etal-2024-assessing}. Videos were segmented using PySceneDetect\footnote{\url{https://github.com/Breakthrough/PySceneDetect}} at natural scene transitions.

\paragraph{Audio Description as Event Source.}
We used transcribed movie audio descriptions to detect events, following prior work \cite{7298940, 10944028,han2023autoad,han2023autoad2,han2024autoad3} that highlights the importance of audio descriptions as high-quality, human-created visual narratives. Movie audio descriptions are explicitly written to describe on-screen visual content for visually impaired audiences, making them reliable gold captions that closely align with visual events and actions \cite{7298940}. Recent studies further emphasize the narrative richness and temporal coherence of audio descriptions for modeling complex event structures in movies, both in automatic generation settings and downstream understanding tasks \cite{10944028,han2023autoad,han2023autoad2,han2024autoad3}. Leveraging these human-authored descriptions allows us to ground event extraction in semantically precise and visually faithful textual representations, improving the quality of extracted triggers, arguments, and event relations.

\subsection{Event Annotation Pipeline}

\paragraph{Event Trigger Detection.}
We extracted event triggers, which indicate when events occur and are denoted as $t_i$ for event $e_i$, using a combination of supervised models and LLM-based methods. In particular, we employed OmniEvent \cite{peng-etal-2023-omnievent} for open-domain trigger detection and augmented it with LLM-based extraction constrained to a predefined predicate set. Following VidSitu \cite{9577635}, we adopted PropBank-selected verbs as the trigger vocabulary to ensure consistency across events and reduce spurious detections. On the GOLD subset, approximately 17\% of narrative events have no exact PropBank trigger and in these cases, annotators select the closest available predicate. We retain PropBank because its fixed predicate senses and argument structures provide a consistent ontology for training and evaluation. This setup enabled robust identification of both atomic and narrative-level event triggers across diverse textual inputs.

\paragraph{Event Argument Extraction.}
We extracted event arguments, which capture the participants and contextual details of events and are denoted as $a_{ij}$ for argument $j$ of event $i$, using LLM-based semantic role extraction from captions and scripts. To improve reliability and coverage, we supplemented LLM outputs with predictions from GLEN \cite{li-etal-2023-glen} and OmniEvent \cite{peng-etal-2023-omnievent}. All arguments followed PropBank \cite{palmer-etal-2005-proposition} conventions, including core roles such as ARG0 and ARG1, as well as modifier roles such as ARGM-LOC for location and ARGM-TMP for time, enabling structured and consistent representation of event semantics.

\paragraph{Event Relation Extraction.}
We extracted event relations, which model how events are connected and are denoted as $r(e_i, e_j)$. The relation ontology includes temporal, causal, preconditioned, hierarchical, coreference, and no-relation labels. Temporal relations capture ordering such as before, after, or overlap, which are grouped under the temporal category for evaluation. We identified event relations using LLM-based methods with text-based extraction techniques inspired by prior document-level event extraction work \cite{10.1007/978-3-031-56069-9_38}. This approach allowed us to capture coherent event structures and dependencies across multiple events within complex narratives.

\paragraph{Video Event Localization.}
We localized events to video timestamps $[t_s, t_e]$ using state-of-the-art temporal video grounding models \cite{wang2025timer} and Gemini 2.5 Pro \cite{comanici2025gemini25pushingfrontier}. This problem can also be framed as a temporal video grounding task, where the goal is to align a natural-language event description with its corresponding time segment in the video. To assess their capability, we created a test set with short videos from VidEvent \cite{liang2025videventlargedatasetunderstanding}. These methods struggled with videos of only a few minutes (Appendix~\ref{sec:appendix-localization}). As a conservative fallback, we therefore used the scene boundaries containing each event as its time boundaries, providing a safe, high-recall temporal localization.

\paragraph{Annotation Protocol.}
We constructed a large-scale \textit{SILVER} event annotation dataset by annotating 1{,}005 movies using an LLM-assisted (Grok-4.1 Fast) \cite{xAI_Grok4_1Fast_2025} pipeline. The annotations covered an average of 70 visually observed events, 31 dialogue-content-based events, and 2 audio events per movie, enabling comprehensive multimodal event coverage. During the creation and verification process, we consumed approximately 100 billion tokens. To ensure annotation quality, we employed a two-step verification strategy in which extracted events were jointly validated against both movie plot summaries and movie audio descriptions. In addition to the SILVER dataset, we curated a high-quality GOLD benchmark by fully annotating 5 movies, each containing approximately 70 events and 50 event relations. 

The GOLD annotations were produced by five contracted human annotators, with each movie independently annotated by two annotators. Annotators were trained on our custom annotation platform and compensated at a rate of \$15 per hour. Annotation required approximately 8 total annotator-hours per movie across the two independent passes (approximately 4 hours per pass), for approximately 40 annotator-hours and \$600 in total.
We evaluated annotation consistency using weighted Cohen's $\kappa$ and mean semantic similarity, measuring both inter-annotator agreement on the GOLD set and agreement between GOLD and SILVER annotations (Figure~\ref{fig:metrics_comparison}). Specifically, GOLD-SILVER weighted Cohen's $\kappa$ is approximately 0.50, compared to inter-annotator $\kappa$ of approximately 0.57 on the GOLD set, indicating that the SILVER pipeline tracks human judgments at roughly 86--88\% of observed human-human consistency. This narrow gap suggests the scalable annotation process is aligned with independent human annotation at a level close to the variation between human annotators themselves.

\paragraph{Dataset Statistics and Data Release.}
NEST is partitioned at the movie level into 80/15/5 train/validation/test splits, with no movie overlap across splits. Evaluation is conducted on subsets of the held-out test split whose sizes vary across task--model configurations due to task-specific requirements and model-specific processing constraints (Appendix~\ref{sec:appendix-bootstrap}). We provide pre-extracted \textit{video-level features}, \textit{frame-level features} computed using models~\cite{tschannen2025siglip2multilingualvisionlanguage, pmlr-v139-radford21a, oquab2024dinov}, and \textit{audio features} using models~\cite{3495724.3496768, 10095889}, enabling research without requiring access to the full videos under our user agreement. A subset of full-length movies in the public domain or under permissive open licenses is also provided for end-to-end research without access restrictions.

The downstream tasks shown in Figure~\ref{fig:NEST}, including long-video captioning and video question answering, are supported by NEST narrative information extraction but are not directly evaluated in this work. NEST instead benchmarks structured extraction and reasoning over narrative events.

\section{Experiments}
\subsection{Training Setup}
We fine-tune Qwen3-Omni-30B-A3B-Instruct on full-length movie inputs using 8 NVIDIA H200 GPUs with mixed-precision training. The visual encoder and multimodal alignment layers are frozen, and language model parameters are updated via low-rank adaptation \cite{hu2022lora}, preserving pretrained visual representations while enabling stable optimization under memory constraints. Since full-length movies span 1 to 2 hours and contain many visually redundant frames, we sparsely sample video inputs at 0.1 FPS. We use gradient checkpointing to reduce activation memory and Flash Attention \cite{3600270.3601459} to support the resulting long-context sequences. Additional training details, token budget calculations, and per-baseline sampling configurations are provided in Appendix~\ref{sec:appendix-sampling}.

\subsection{Benchmark Tasks}
\vspace{-0.4em}
We define four complementary tasks over full-length movies. Event trigger detection (ETD) asks models to identify the narrative event given video $V$ and scene boundaries $[t_s, t_e]$. Models return a trigger verb and context that captures the story-level action taking place in the scene, rather than surface-level physical motions. Event localization (EL) is formulated as scene-level narrative grounding: given an event description (verbs and arguments), models predict temporal boundaries $[\hat{t}_s, \hat{t}_e]$ for where that event occurs in the movie, evaluated by checking whether predictions fall within the ground-truth scene boundaries. We adopt scene-level evaluation rather than frame-precise boundaries because narrative event boundaries are inherently subjective, and even state-of-the-art temporal grounding models struggle on short videos (Appendix~\ref{sec:appendix-localization}). With ${\sim}$160 scenes per movie, identifying the correct scene still requires aligning events across a 1--2 hour narrative with long gaps, subplots, and flashbacks. Event argument extraction (EAE) provides a trigger and semantic roles within a scene and asks models to fill in the argument values, such as who performed the action, who was affected, and where it took place.  Event relation extraction (ERE) gives models a pair of events $(e_i, e_j)$ specified by their definitions and asks them to predict the relation $\hat{r}(e_i, e_j) \in \{$temporal, causal, preconditioned, hierarchical, coreference, no relation$\}$, measured by Precision, Recall, and F1 directly against ground-truth labels. Since non-linear temporal structure is common in movies, we also evaluate models on flashback detection as a separate subset of ERE.

\paragraph{Text-Only Narrative Event Extraction.}
In addition to the video-based tasks above, we evaluate a text-only variant to isolate language-only performance. Here we replace video input with captions produced by Gemini~2.5~Pro \cite{comanici2025gemini25pushingfrontier} for each scene and perform event extraction purely from text. We treat audio descriptions as a privileged modality since they are not consistently available across videos, and instead use automatically generated captions to provide a uniformly applicable text-only baseline. We report Precision, Recall, and F1@k against the same ground-truth events (Table \ref{tab:event-extraction}).

\subsection{Baselines}

We evaluate NEST using a range of state-of-the-art long video understanding and multimodal models shown in Table~\ref{tab:event-understanding}, including Qwen3-VL \cite{bai2025qwen3vltechnicalreport}, Qwen3-Omni \cite{xu2025qwen3omnitechnicalreport}, Qwen2.5-VL \cite{bai2025qwen25vltechnicalreport}, InternVL3.5 \cite{wang2025internvl35advancingopensourcemultimodal}, Video-LLaMA3 \cite{zhang2025videollama3frontiermultimodal}, LLaVA-Video \cite{zhang2025llavavideo}, Ovis2.5 \cite{lu2025ovis25technicalreport}, Flash-VStream-Qwen \cite{zhang2025flashvstreamefficientrealtimeunderstanding}, and LongVU \cite{pmlr-v267-shen25j}. These models are evaluated on multiple complementary tasks: (1) event trigger detection (ETD), (2) event localization (EL), (3) event argument extraction (EAE), (4) event relation extraction (ERE) across relation types, and (5) flashback relation identification, which isolates non-linear temporal reasoning beyond linear event timelines.

\begin{table*}[!t]
\centering
\scriptsize
\resizebox{\textwidth}{!}{
\begin{tabular}{lcccccccccccc}
\toprule
\multirow{2}{*}{Method} &
\multirow{2}{*}{\#Params} &
\multirow{2}{*}{Video Sampling} &
\multicolumn{1}{c}{ETD} &
\multicolumn{1}{c}{EL} &
\multicolumn{1}{c}{EAE} &
\multicolumn{7}{c}{ERE -- F1 (\%)} \\
\cmidrule(lr){4-4}
\cmidrule(lr){5-5}
\cmidrule(lr){6-6}
\cmidrule(lr){7-13}
 & & &
Acc (\%) &
Acc (\%) &
Acc (\%) &
no\_rel & 
coref &
hier &
precond &
temp &
causal &
overall \\
\hline
\rowcolor{gray!5} \multicolumn{13}{c}{\textit{Zero-shot 1fps models}} \\
\hline
Qwen3-VL (8B) \cite{bai2025qwen3vltechnicalreport} & 8B & 1 FPS & 3.42 & 0.87 & 3.03 & 39.61 & 35.68 & 0.00 & 0.00 & 0.38 & 31.34 & 17.84 \\

Qwen3-VL (30B) \cite{bai2025qwen3vltechnicalreport} & 30B & 1 FPS & 3.48 & \textbf{5.89} & 4.60 & 37.52 & 55.46 & 42.11 & 0.00 & 0.00 & 14.90 & 25.00 \\

Qwen3-Omni \cite{xu2025qwen3omnitechnicalreport} & 30B & 1 FPS & 3.20 & 0.44 & 7.40 & 8.99 & 26.67 & 8.22 & 11.76 & \textbf{14.29} & 44.87 & 19.13 \\

Qwen2.5-VL (7B) \cite{bai2025qwen25vltechnicalreport} & 7B & 1 FPS & 4.33 & 0.66 & 3.93 & 26.56 & 41.38 & 0.00 & 0.00 & 4.22 & 8.31 & 13.41 \\

Qwen2.5-VL (32B) \cite{bai2025qwen25vltechnicalreport} & 32B & 1 FPS & 1.67 & 0.26 & 3.38 & \textbf{42.79} & 61.95 & 50.52 & 5.61 & 10.21 & 34.31 & 34.23 \\

LongVU-LLaMA3 \cite{pmlr-v267-shen25j} & 3B & 1 FPS & 1.38 & 0.61 & 0.31 & 5.39 & 0.00 & 2.66 & 5.05 & 0.00 & 31.79 & 7.48 \\

LongVU-Qwen2 \cite{pmlr-v267-shen25j} & 7B & 1 FPS & 0.49 & 0.41 & 1.55 & 16.96 & 0.00 & 2.50 & 0.00 & 0.00 & 34.28 & 8.96 \\

Video-LLaMA3 \cite{zhang2025videollama3frontiermultimodal} & 7B & 1 FPS & 2.76 & 0.92 & 0.00 & 37.50 & 26.32 & 8.33 & 6.61 & 13.71 & 27.50 & 20.00 \\

\hline
\rowcolor{gray!5} \multicolumn{13}{c}{\textit{Zero-shot frame-selection models}} \\
\hline
Ovis2.5 \cite{lu2025ovis25technicalreport} & 9B & 8 frames & 7.27 & 0.00 & \textbf{10.62} & 37.35 & 28.57 & 0.00 & 0.00 & 5.45 & 24.10 & 15.91 \\ 

InternVL3.5 \cite{wang2025internvl35advancingopensourcemultimodal} & 30B & 32 frames & 2.34 & 0.53 & 2.89 & 37.50 & 56.07 & 22.12 & 4.04 & 4.79 & 18.52 & 23.84 \\

LLaVA-Video \cite{zhang2025llavavideo} & 7B & 64 frames & \textbf{7.98} & 0.33 & 10.25 & 4.62 & 0.00 & 10.60 & 0.00 & 0.00 & 37.71 & 8.82 \\ 

\hline
\rowcolor{gray!5} \multicolumn{13}{c}{\textit{Zero-shot online streaming models}} \\
\hline
Flash-VStream-Qwen \cite{zhang2025flashvstreamefficientrealtimeunderstanding} & 7B & 1 FPS (stream) & 3.98 & 0.53 & 1.25 & 35.22 & 15.38 & 8.57 & 0.00 & 0.00 & 10.67 & 11.64 \\

\hline
\rowcolor{gray!5} \multicolumn{13}{c}{\textit{Finetuned}} \\
\hline
Finetuned Qwen3-Omni (Ours) & 30B & 1 FPS & 6.09 & 0.45 & 10.50 & 15.58 & \textbf{100.00} & \textbf{76.82} & \textbf{38.26} & 2.44 & \textbf{62.12} & \textbf{49.20} \\
\bottomrule
\end{tabular}
}
\caption{Performance across narrative understanding tasks, including video narrative event trigger detection (ETD), video narrative event localization (EL), video narrative event argument extraction (EAE), and video narrative event relation extraction (ERE). Evaluation set sizes vary across task--model configurations because full-length movie inference is subject to model-specific context, frame-budget, and computational constraints. Selected headline configurations and their movie counts are reported in Appendix~\ref{sec:appendix-bootstrap}. Video Sampling denotes either the sampling rate or the fixed frame budget used by each model. The fine-tuned Qwen3-Omni model is trained at 0.1 FPS and evaluated at 1 FPS for the primary results, with matched 0.1-FPS results reported in the analysis.}
\label{tab:event-understanding}
\end{table*}

\begin{table}[t]
\centering
\scriptsize
\resizebox{\columnwidth}{!}{
\begin{tabular}{lccc}
\toprule
Model & Precision & Recall & F1 \\
\midrule
Gemini 2.5 Pro \cite{comanici2025gemini25pushingfrontier} \textsuperscript{*} & 22.58 & 34.18 & 21.09  \\
GPT-5 \cite{singh2025openaigpt5card} \textsuperscript{*} & 40.95 & 44.64 & \textbf{40.80} \\
\midrule
Qwen3-VL (8B) \cite{bai2025qwen3vltechnicalreport} & 18.39 & 44.74 & 17.84 \\
Qwen3-VL (30B) \cite{bai2025qwen3vltechnicalreport} & 31.06 & 48.08 & 25.00 \\
Qwen3-Omni \cite{xu2025qwen3omnitechnicalreport} & 23.19 & 37.90 & 19.13 \\
Qwen2.5-VL (7B) \cite{bai2025qwen25vltechnicalreport} &  18.05 & 33.96 & 13.41 \\
Qwen2.5-VL (32B) \cite{bai2025qwen25vltechnicalreport} & \textbf{42.31} & \textbf{50.54} & 34.23 \\
LongVU-LLaMA3 \cite{pmlr-v267-shen25j} & 9.55 & 11.05 & 7.48 \\
LongVU-Qwen2 \cite{pmlr-v267-shen25j} & 10.31 & 16.10 & 8.96 \\
Video-LLaMA3 \cite{zhang2025videollama3frontiermultimodal} & 23.82 & 32.84 & 20.00 \\
Ovis2.5 \cite{lu2025ovis25technicalreport} & 20.07 & 42.60 & 15.91 \\ 
InternVL3.5 \cite{wang2025internvl35advancingopensourcemultimodal} & 31.63 & 43.80 & 23.84 \\
LLaVA-Video \cite{zhang2025llavavideo} & 15.51 & 15.49 & 8.82 \\ 
Flash-VStream-Qwen \cite{zhang2025flashvstreamefficientrealtimeunderstanding} & 15.84 & 32.26 & 11.64 \\
\bottomrule
\end{tabular}
}
\caption{Video Narrative Event Relation Extraction (ERE) zero-shot performance. F1 is macro-averaged over the six relation types, while Precision and Recall are reported at the prediction level. \textsuperscript{*}Gemini 2.5 Pro and GPT-5 are evaluated on 10 videos at an API cost of approximately \$500, so these results are not directly comparable to the other results in Table~\ref{tab:event-understanding}. See Table~\ref{tab:gpt5-nest-sampled} for GPT-5 per-task results.}
\vspace{-2em}
\label{tab:event-relation}
\end{table}

\subsection{Evaluation Metrics}

We match the evaluation method to the nature of each task's output. ERE predictions come from a closed set of six relation types, so we compute F1 directly without any judge, reporting both overall and per-type scores. EL compares predicted timestamps against ground-truth scene boundaries via automatic overlap, and we additionally report flashback-subset F1 for non-linear temporal reasoning. ETD and EAE produce open-ended outputs where exact matching is too brittle. A model predicting ``fight'' for ground-truth ``attack'' has identified the correct event, and the same character may appear as ``the detective'' or ``Officer Miller'' across different models. For these two tasks, we use an LLM-based judge to assess semantic equivalence. We further evaluate cross-judge robustness using an independent model-family judge on a stratified sample of approximately 200 ETD/EAE predictions, obtaining 0.88 agreement (Table~\ref{tab:judge-validation}). Full judge details and prompts are provided in Appendix~\ref{sec:appendix-judge}.

\section{Results and Analysis}

NEST is designed to test two distinct capabilities that are often conflated in long-video work: (i) \emph{grounded narrative event discovery} (detecting what happens, where it happens, and who/what is involved), and (ii) \emph{reasoning over an event graph} once candidate events are specified. 
Tables~\ref{tab:event-understanding} -- \ref{tab:event-extraction} together show that current long-video models struggle primarily with the first capability, and only partially succeed at the second.

\textbf{Chance baselines establish that low scores reflect task difficulty, not evaluation breakage.} EL is a selection problem with ${\sim}160$ scene candidates per movie, giving a random chance floor of ${\sim}0.6\%$. The best zero-shot model achieves $5.89\%$ (Qwen3-VL 30B), approximately $9\times$ random. ETD and EAE are evaluated under a permissive synonym-tolerant LLM judge against the PropBank vocabulary, where paraphrases are accepted as correct. Under this forgiving setup, the best scores remain below $8\%$ and $11\%$. ERE reaches $34.23\%$ F1 zero-shot under direct closed-set scoring, showing that relation classification given events is substantially more tractable than grounded event discovery. Together these reference points show that grounded narrative event discovery is genuinely hard for current models, while relation classification given events is more tractable.

\findingz{1}{Grounded narrative event discovery remains largely unsolved. ETD stays below 8\%, EL below 6\%, and EAE below 11\%. Higher frame budgets do not consistently help, indicating that visual coverage alone is insufficient.}

\begin{table}[t]
\centering
\scriptsize
\resizebox{\linewidth}{!}{%
\begin{tabular}{lccc|ccc}
\toprule
Model & P@5 & R@5 & F1@5 & P@10 & R@10 & F1@10  \\
\midrule
\multicolumn{7}{l}{\textit{Vision-Language Models}} \\
\midrule
Gemini 2.5 Pro \cite{comanici2025gemini25pushingfrontier} & 4.2 & 7.9 & 5.2  & 4.2 & 7.9 & 5.2  \\
\midrule
\multicolumn{7}{l}{\textit{Text Only EE Methods}} \\
\midrule
GLEN \cite{li-etal-2023-glen} & 5.6 & 15.3 & \textbf{8.2} & 5.9 & 17.6 & \textbf{8.8} \\
OmniEvent \cite{peng-etal-2023-omnievent} & 6.7 & 5.4 & 6.0 & 6.8 & 5.9 & 6.3 \\
\bottomrule
\end{tabular}
}
\caption{Video Narrative Event Extraction performance measured by Precision (P), Recall (R), and F1 score @ [5, 10]. All models are evaluated using only the scene captions text as input. Evaluation is conducted on 337 scenes across 10 videos.}
\vspace{-2em}
\label{tab:event-extraction}
\end{table}

\begin{figure*}[t]
    \centering
    \includegraphics[width=\linewidth]{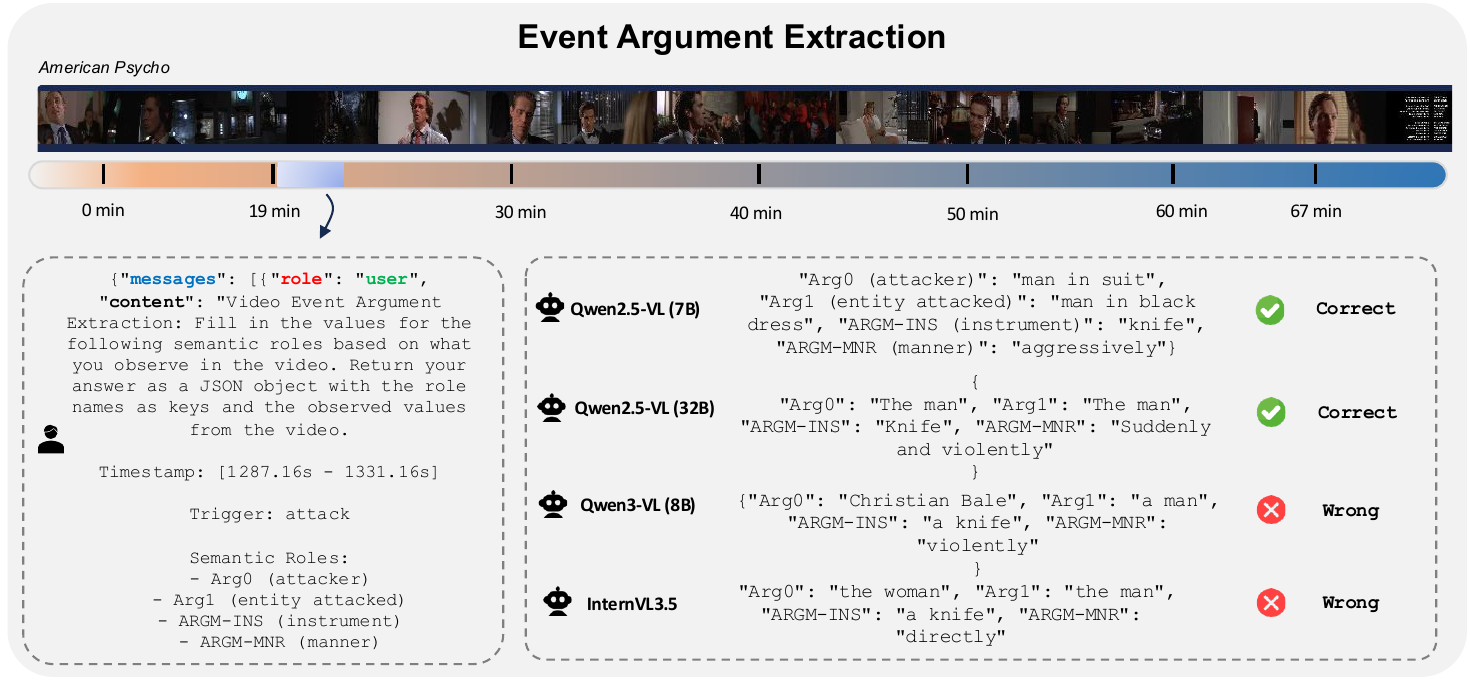}
    \caption{Event Argument Extraction example from \textit{American Psycho} (102 min). Given the ``attack'' event trigger and four semantic roles (ARG0: attacker, ARG1: entity attacked, ARGM-INS: instrument, ARGM-MNR: manner), models extract the corresponding argument values. Both Qwen2.5-VL variants (7B and 32B) correctly identify the arguments. Qwen3-VL (8B) hallucinates the actor name ``Christian Bale'' as the attacker instead of describing the character from the video. InternVL3.5 hallucinates a woman as the attacker even though the scene contains two male participants, while also incorrectly assigning the attacked entity.}
    \label{fig:eae-example-main}
\end{figure*}

\textbf{Event argument extraction (EAE)} (Table~\ref{tab:event-understanding})
remains below 11\% accuracy for all methods. 
Even when a correct trigger is plausible, models often fail to 
recover structured information such as participants, roles, and attributes. 
Figure~\ref{fig:eae-example-main} illustrates these failures, including incorrect entity identification and semantic role assignment.
This gap is particularly important for narrative understanding because downstream relation links depend on consistent entities and roles across distant scenes.

Our error analysis further shows that this failure is driven primarily by entity tracking rather than missing argument predictions (Appendix~\ref{sec:appendix-qualitative}).

\findingz{2}{EAE is dominated by entity confusion. Across five models, 97.6\% of errors involve entity identity, while only 2.4\% are missing arguments, highlighting the need for long-range identity tracking across long videos.}

\textbf{Event relation extraction (ERE)} (Table~\ref{tab:event-understanding}) 
is substantially easier when events are provided, but temporal structure remains weak.
Given there are a fixed number of relations to classify between, this task is easier for models. 
Our results show that models sometimes reason about relations \emph{conditional on having event descriptions}, but they struggle to discover and ground those events from raw movies in the first place.
Zero-shot ERE peaks at \textbf{34.23\% F1} (Qwen2.5-VL 32B \cite{bai2025qwen25vltechnicalreport}), and our fine-tuned model improves overall ERE to \textbf{49.20\% F1}. 
Nevertheless, performance varies sharply by relation type. Coreference and (to a lesser extent) causal relations are learned more reliably, while temporal, precondition, and hierarchical relations remain challenging for many models. The particularly high fine-tuned scores for coreference (100.00 F1) and hierarchical relations (76.82 F1) should be interpreted in light of their relatively small evaluation support, with 37 and 139 instances, respectively. This is consistent with the long-range narrative dependencies in NEST (Figure~\ref{fig:Temporal distance between}), where related events can be separated by large temporal gaps and intervening scenes.

\findingz{3}{Models are confidently wrong at localization but often conservative at relation extraction. Across five models, 97.4--99.4\% of EL errors are wrong-scene predictions, while four of five models predict NO\_RELATION for over half of ground-truth related pairs.}

\paragraph{Fine-tuning helps reasoning more than grounding.}
Fine-tuning Qwen3-Omni on NEST substantially improves ERE (49.20\% F1 overall; Table~\ref{tab:event-understanding}), but does not yield comparable gains in ETD/EL. To disentangle supervision gains from sampling density, we additionally evaluate zero-shot and fine-tuned Qwen3-Omni under the same 0.1-FPS sampling rate. At 0.1 FPS, the zero-shot model obtains 2.83\% ETD, 0.38\% EL, and 34.61\% ERE F1, while the fine-tuned model obtains 4.15\%, 0.39\%, and 37.50\%, respectively. Thus, under matched sampling, fine-tuning improves ERE by 2.89 points while EL remains essentially unchanged (+0.01) and ETD improves only modestly (+1.32). This control shows that the qualitative separation between conditional relation reasoning and grounded event discovery is not solely explained by sampling density. ERE is a conditional closed-set task where both events are already specified, whereas ETD and EL require the model to search over the full movie, infer story-level abstractions from low-level multimodal evidence, and ground them to the correct scene among ${\sim}$160 candidates. Our error analysis (Appendix~\ref{sec:appendix-qualitative}, Tables~\ref{tab:error-taxonomy-models}--\ref{tab:error-taxonomy}) supports this reading: across the five analyzed models, ETD failures are dominated by wrong narrative events (76.4\% on average), EL failures are almost entirely wrong-scene predictions (98.4\%), and EAE failures are overwhelmingly entity confusions (97.6\%). These patterns indicate that fine-tuning can improve the decision rule once the event abstraction is provided, but does not by itself give the model robust long-range visual memory, entity tracking, or narrative abstraction.

\textbf{Flashback relations.}
To further probe non-linear temporal reasoning, we evaluate models on the NEST flashback subset (Table~\ref{tab:f1_flashback_subset}), which isolates event relations that violate linear chronological order. Performance drops catastrophically: five of seven models achieve 0.00 F1, with only Qwen2.5-VL (7B and 32B) \cite{bai2025qwen25vltechnicalreport} achieving non-zero performance, at 13.04\% and 28.57\% F1, respectively. Notably, larger models including InternVL3.5 (30B) \cite{wang2025internvl35advancingopensourcemultimodal} and both Qwen3-VL \cite{bai2025qwen3vltechnicalreport} variants completely fail despite stronger performance on general event relation extraction. These results suggest that non-linear temporal reasoning poses a distinct challenge beyond standard temporal ordering, as current models rely heavily on implicit linear timeline assumptions, limiting their ability to handle narrative constructs such as flashbacks, memory recall, etc., common in real-world videos.

\findingz{4}{Relations involving flashbacks are especially challenging. Five of seven models achieve 0.00 F1 on the flashback subset, showing difficulty linking events embedded in non-linear narrative structure.}

\section{Conclusion}

We introduce NEST, a benchmark for narrative event understanding in full-length movies with structured annotations, temporal boundaries, and inter-event relations across 1,005 films grounded in visual content, dialogue, and audio. State-of-the-art models struggle across all tasks, revealing that long video processing does not yield narrative-level comprehension. Results highlight the need for narrative abstraction, temporal reasoning, and hierarchical understanding beyond frame sampling.

\section{Limitations}

Our annotation pipeline uses LLMs for verifying detected narrative events, which may introduce errors or biases inherent to these models. The use of a PropBank-constrained trigger vocabulary also limits ontology coverage as approximately 17\% of GOLD narrative events do not have an exact PropBank trigger and must be represented using the closest available predicate. The GOLD subset contains only five full-length movies, which limits the statistical power of evaluation despite comprising approximately 350 events and 250 relation annotations; results on GOLD should therefore be interpreted with this sample size in mind. While we use state-of-the-art temporal grounding models for event localization, these methods struggle even on short videos (see Appendix~\ref{sec:appendix-localization}), which is why we evaluate at scene-level rather than frame-precise temporal boundaries. This scene-level evaluation may not capture fine-grained temporal understanding. Additionally, NEST annotates events within individual scenes, which means that events that can only be detected by jointly reasoning across multiple scenes (e.g., a gradual character transformation or a subplot that unfolds over several non-adjacent scenes) are not represented in the current annotation. Capturing such cross-scene composite events is an important direction for future work. Although Figure~\ref{fig:NEST} illustrates captioning and video question answering as potential downstream applications of NEST's narrative representations, we do not evaluate transfer to these tasks in this work. The NEST corpus is skewed toward older, predominantly Western films, with uneven genre and language coverage, which may limit generalizability to contemporary, non-Western, and linguistically diverse video. Due to copyright restrictions, we cannot release raw videos for the full NEST corpus; instead, we release pre-extracted video-level, frame-level, and audio features, together with the raw full-length videos for a subset of movies that are in the public domain or available under permissive open licenses. This restriction may limit certain types of analysis or model development on the remaining copyrighted movies. Finally, our LLM-as-a-judge evaluation for event trigger detection and event argument extraction, while necessary for semantic matching, introduces potential biases from the judge model's own limitations.

\section{Ethical Considerations}

NEST uses publicly available movies from the Library of Congress, archive.org, PublicDomainMovies, YouTube, and other databases under fair use for academic research. We do not redistribute copyrighted raw videos. For non-public-domain content we release only pre-extracted features under a research-only agreement. Additionally we release a subset of public-domain/permissively licensed movies for fully reproducible experiments. Movies may contain biases, stereotypes, or problematic representations from their time of creation, which models may inadvertently learn. Released multimodal features and structured coreference annotations could also be repurposed for unintended analyses beyond the intended research use. We encourage responsible use of NEST solely for advancing narrative understanding research and discourage applications enabling surveillance, manipulation, or other harmful uses.

\section*{Acknowledgments}

We acknowledge Advanced Research Computing (ARC) at Virginia Tech for providing computational resources and technical support that contributed to the results reported in this paper. We also thank the anonymous reviewers for their constructive feedback, which helped improve the paper. Generative AI assistants were used for language editing, while all research design, methodological, experimental, and implementation decisions were made and verified by the authors.

\bibliography{custom}
\appendix
\section{Additional Experimental and Dataset Details}
\label{sec:appendix}
This appendix provides additional details and analyses that complement the main paper. We first present the NEST dataset distribution (Appendix~\ref{sec:appendix-distribution}) and further motivate our scene-level localization setup (Appendix~\ref{sec:appendix-localization}). We then provide reproducibility and release details (Appendix~\ref{sec:appendix-release}), followed by additional analysis of annotation quality and human annotation effort (Appendix~\ref{sec:appendix-annotation-quality}). We also describe the video sampling and context-handling configurations used across models (Appendix~\ref{sec:appendix-sampling}), report bootstrap confidence intervals (Appendix~\ref{sec:appendix-bootstrap}), and provide details of the LLM-as-a-judge evaluation (Appendix~\ref{sec:appendix-judge}). Finally, we present qualitative examples and error analyses (Appendix~\ref{sec:appendix-qualitative}), discuss directions for future work (Appendix~\ref{sec:appendix-future}), and provide the annotation platform (Appendix~\ref{sec:appendix-platform-ui}), annotator instructions (Appendix~\ref{sec:appendix-platform-prompts}), and task-specific prompt templates (Appendix~\ref{sec:appendix-task-definitions}).
\subsection{Dataset Distribution}
\label{sec:appendix-distribution}
\begin{figure}
    \centering
    \includegraphics[width=1\linewidth]{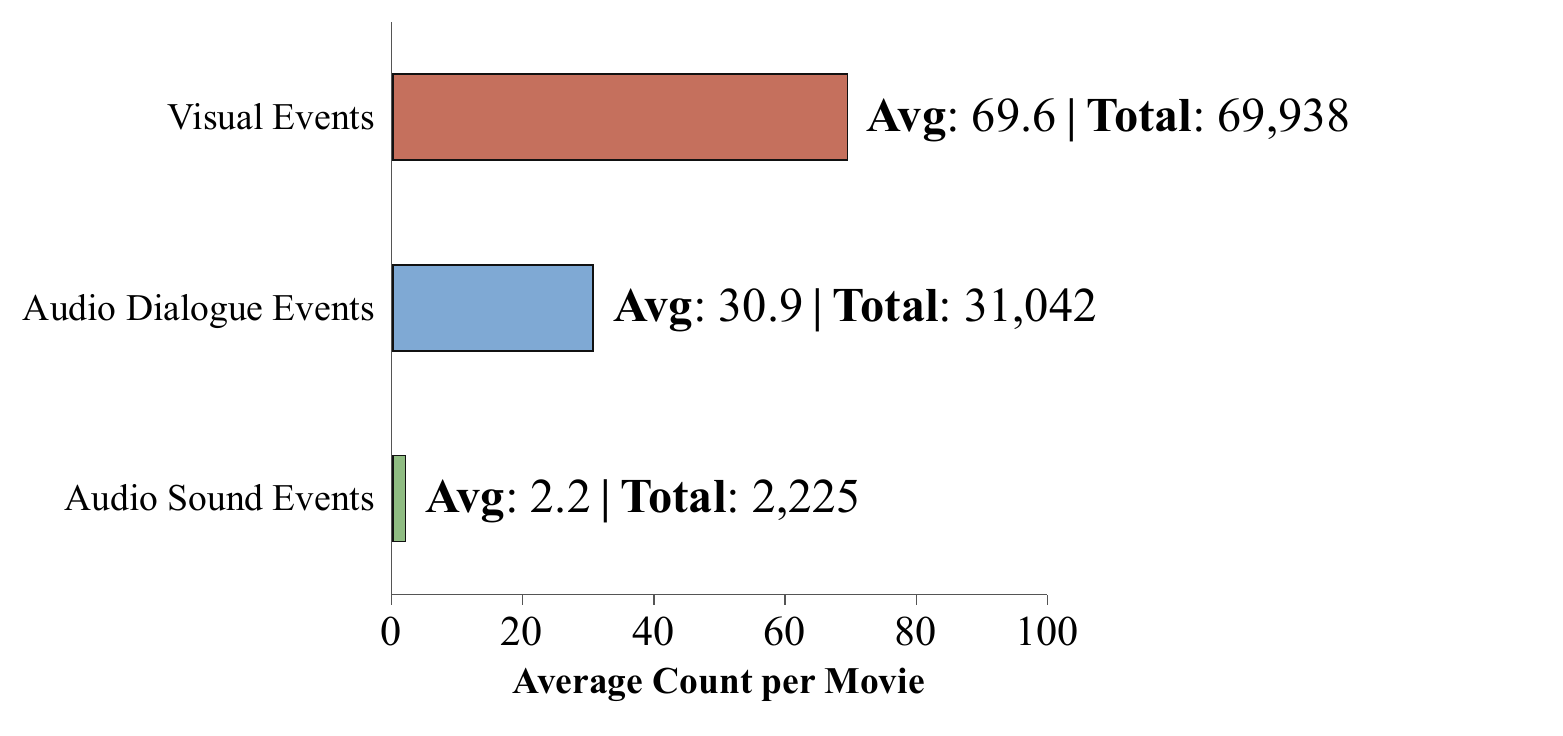}
    \caption{Average number of annotated events per movie in the NEST dataset, broken down by visual events, audio dialogue events, and audio sound events. Bars show per-movie averages, with total counts reported for each modality.}
    \label{fig:Average number of}
    \vspace{-1.5em}
\end{figure}

Figure~\ref{fig:Average number of} shows the distribution of annotated events across modalities in NEST. Averaged over all 1,005 movies, NEST contains 69.6 visual events per movie (69,938 total), 30.9 dialogue-based events per movie (31,042 total), and 2.2 audio sound events per movie (2,225 total). Together, these distributions reflect the multimodal nature of NEST while showing that most narrative events are grounded primarily in visual content and dialogue.

\subsection{Event Localization Rationale}
\label{sec:appendix-localization}

We adopt scene-level localization rather than fine-grained temporal boundaries for two reasons.
First, the boundaries of narrative events are inherently subjective.
An event such as ``breakup'' may arguably begin as tension escalates, include several decisive moments, and resolve as characters separate. Different annotators can reasonably place start and end times at different moments while fully agreeing on the event and its approximate location in the narrative.
Second, even state-of-the-art temporal grounding models struggle to produce reliable boundaries on short videos, as shown in Table~\ref{tab:our-results}.

\begin{table}[!ht]
\centering
\scriptsize
\resizebox{\columnwidth}{!}{%
\begin{tabular}{lcccc}
\toprule
Model & R@0.3 & R@0.5 & R@0.7 & mIoU \\
\midrule
Gemini 2.5 Flash \cite{comanici2025gemini25pushingfrontier} & 11.68 & 6.04 & 2.39 & 7.87 \\
Gemini 2.5 Pro \cite{comanici2025gemini25pushingfrontier} & 11.48 & 6.07 & 2.33 & 7.96 \\
Time-R1 \cite{wang2025timer} & 36.78 & 18.45 & 6.21 & 23.23 \\
\bottomrule
\end{tabular}}
\caption{Event localization on the evaluation set of \textbf{VidEvent} \cite{liang2025videventlargedatasetunderstanding} remains challenging. Despite operating under relatively small temporal context windows, state-of-the-art temporal grounding models still struggle to accurately localize event boundaries.}
\label{tab:our-results}
\end{table}

The average movie in NEST contains approximately 160 scenes.
Selecting the correct scene (or set of scenes) for a narrative event still requires aligning the event to the correct part of a 1--3 hour multimodal narrative, across long temporal gaps, subplots, and non-linear devices such as flashbacks.
We therefore view scene-level localization as a meaningful and less subjective proxy for temporal grounding at movie scale.
As fine-grained temporal grounding methods improve, NEST can naturally accommodate finer-grained evaluation.

\subsection{Reproducibility and Release Details}
\label{sec:appendix-release}

Table~\ref{tab:release-artifacts} summarizes every artifact that will be released alongside this paper.

\begin{table}[ht]
\centering
\small
\begin{tabular}{lcc}
\toprule
\textbf{Artifact} & \textbf{Release} & \textbf{Format} \\
\midrule
Train / Val / Test splits          & \cmark & JSON \\
SILVER event annotations           & \cmark & JSON \\
GOLD event annotations             & \cmark & JSON \\
Pre-extracted video features       & \cmark & .npy / .pt \\
Pre-extracted audio features       & \cmark & .npy / .pt \\
Feature extraction specifications  & \cmark & YAML \\
Evaluation scripts                 & \cmark & Python \\
Training configurations            & \cmark & YAML \\
LLM-as-a-judge prompts             & \cmark & Text \\
Public-domain movie subset         & \cmark & Video \\
Fine-tuned model checkpoint        & \cmark & .pt \\
\bottomrule
\end{tabular}
\caption{Artifacts provided with NEST to support reproducibility.}
\label{tab:release-artifacts}
\end{table}

We will release a fixed, frozen snapshot of the SILVER annotations, but we also note that the inability to redistribute copyrighted raw movies is not unique to NEST. Several widely used video benchmarks follow the same distribution model, providing pre-extracted features, annotations, and evaluation tooling rather than raw video files. MAD~\cite{soldan2022madscalabledatasetlanguage}, the closest precedent to NEST in scale and domain, explicitly states that due to copyright restrictions raw movies will not be released, and instead provides pre-extracted CLIP frame-level features at 5 FPS along with language token embeddings. MovieQA~\cite{7780870} similarly does not distribute raw movie videos, instead releasing plot synopses, subtitles, and video clips sourced through external links. Other video understanding datasets adopt comparable strategies. HiREST~\cite{10204095} releases pre-extracted EVA-CLIP features and ASR transcripts without raw videos. ReXTime \cite{chen2024rextimebenchmarksuitereasoningacrosstime} releases QA pairs and temporal spans without raw videos. VidChapters-7M~\cite{yang2023vidchaptersm} releases ASR transcripts, chapter boundaries, and titles via video IDs, requiring users to download raw videos independently from YouTube. NEST follows this established convention and additionally releases pre-extracted multimodal features covering several recent vision-language models to further lower the barrier to entry. We will also release a subset of full-length movies that are in the public domain or available under permissive open licenses, enabling end-to-end evaluation without access restrictions.

\subsection{Annotation Quality and Human Effort}
\label{sec:appendix-annotation-quality}

\paragraph{Grounding in Human-Authored Sources.}
The SILVER annotations are not generated from free-form model captions.
The primary textual source is professional audio descriptions written for visually impaired audiences, which are specifically designed to faithfully describe on-screen visual content~\cite{7298940}.
These are supplemented by human-written scripts and plot summaries.
The LLM-assisted pipeline extracts structured events from these human-authored sources, constrained to PropBank's ontology and a closed relation label set, rather than generating events from scratch.
This design is intended to prevent the benchmark from collapsing into ``hallucination matching,'' where labels exist only in text but have no visual correspondence.

\begin{figure}
    \centering
    \includegraphics[width=1\linewidth]{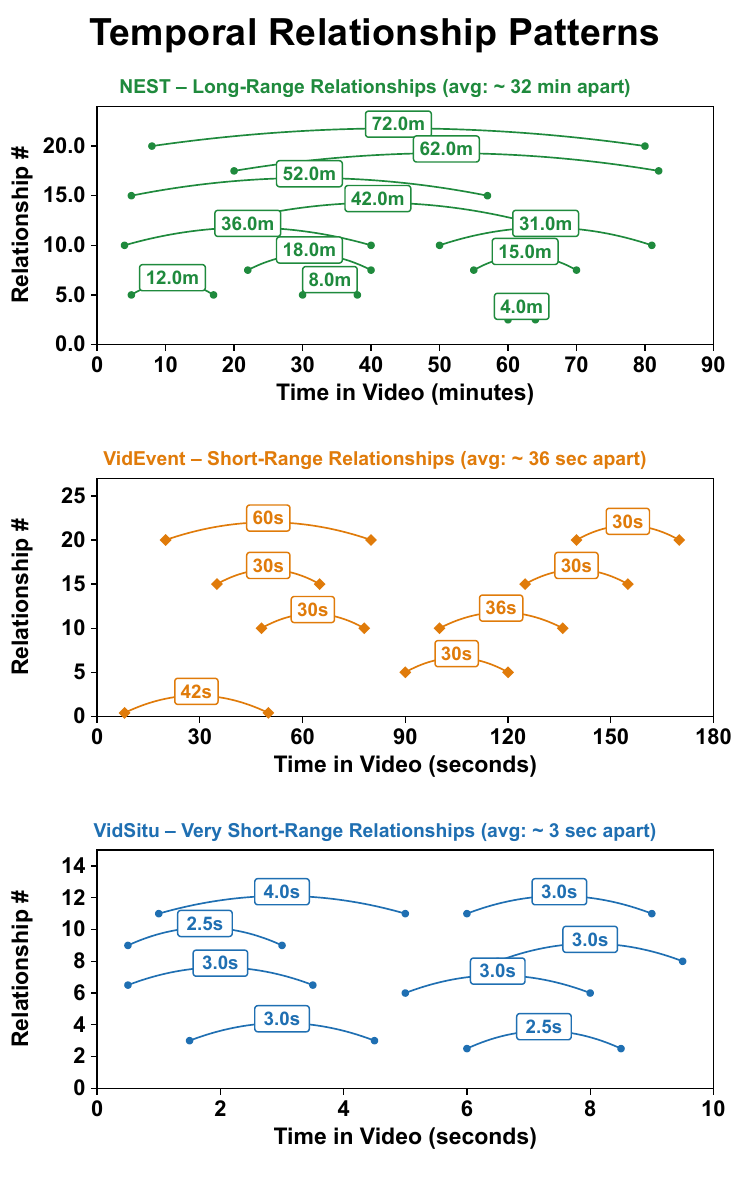}
    \caption{Temporal distance between related events across datasets. Each arc connects a pair of related events, with arc length indicating their temporal separation. NEST captures long-range narrative dependencies averaging ${\sim}32$ minutes apart, compared with ${\sim}36$ seconds for VidEvent and ${\sim}3$ seconds for VidSitu.}
    \label{fig:Temporal distance between}
    \vspace{-1em}
\end{figure}

\paragraph{Two-Step Verification.}
To minimize hallucination risk, extracted events undergo a two-step verification against two independent human-authored signals, each designed to filter different types of errors.

\textit{Step 1: Audio description verification (local visual grounding).}
Each extracted event is checked against the movie's audio description to ensure it corresponds to observable on-screen content. This step filters hallucinated or visually unsupported events. For example, an extracted ``betray'' event was rejected when the audio description for the relevant scene described only two characters having a calm conversation with no indication of deception or betrayal. Similarly, a ``chase'' event was removed when the audio description mentioned characters walking together rather than pursuing one another.

\textit{Step 2: Plot/script verification (global narrative consistency).}
Events passing Step~1 are cross-validated against plot summaries and scripts to verify narrative consistency. This step filters wrong PropBank senses, incorrect arguments, and over-inferred relations. For example, a ``kill'' event extracted from a scene depicting a heated argument was rejected because the plot summary confirmed the character survived. A ``causal'' relation between two events was downgraded to ``temporal'' because the plot indicated the events occurred independently in parallel subplots.

Events that cannot be supported by either source are filtered out. This verification pipeline used GPT-5, GPT-5-nano, and GPT-5-mini~\cite{singh2025openaigpt5card} alongside Grok-4.1 Fast~\cite{xAI_Grok4_1Fast_2025} across both stages. During this creation and verification process, we consumed approximately 100 billion tokens.

\paragraph{Independent Gold Annotation.}
The GOLD annotations were produced independently of the SILVER data.
Five contracted annotators watched full-length movies and produced event and relation annotations from scratch, without access to the SILVER labels.
Agreement between GOLD and SILVER was then measured to validate the automatic pipeline, rather than using the GOLD set as a correction layer on the SILVER data.

\begin{figure}
    \centering
    \includegraphics[width=1\linewidth]
    {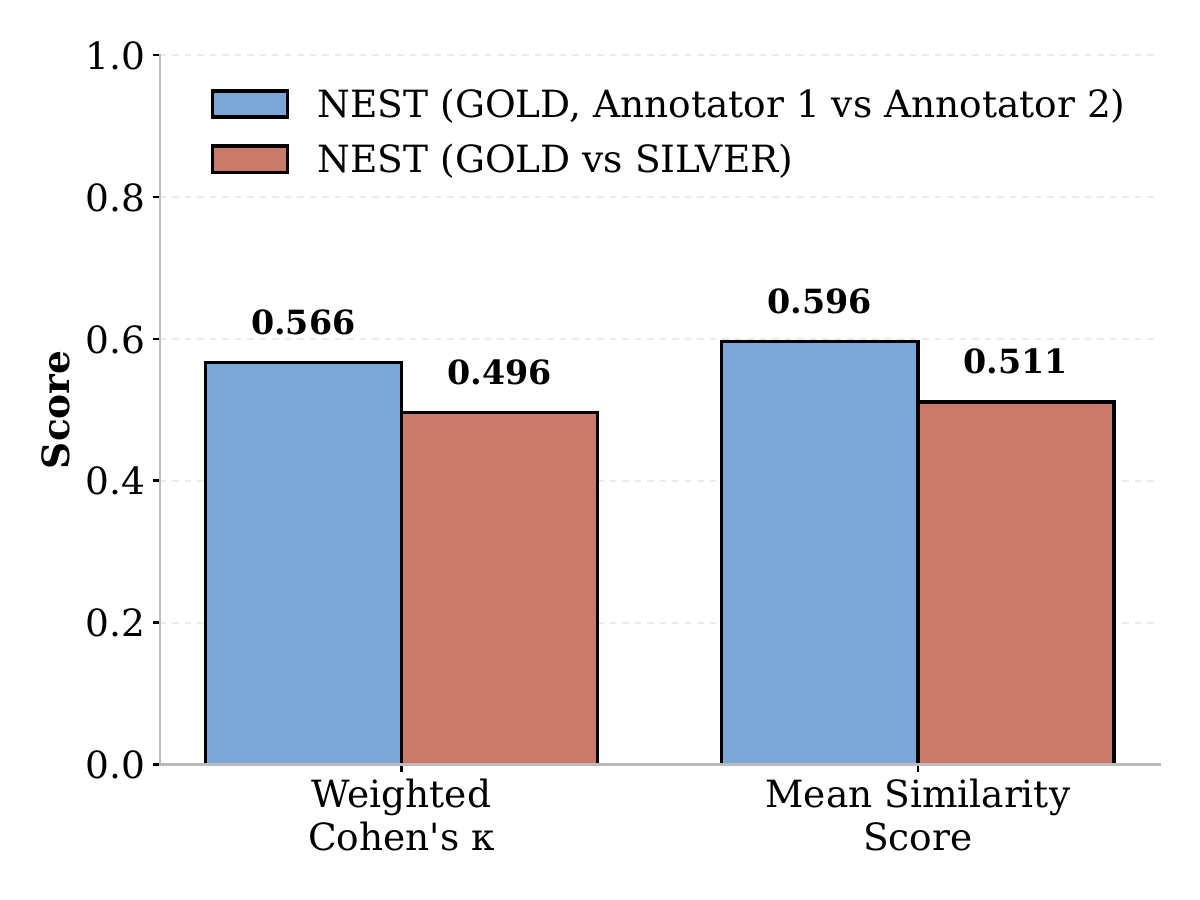}
  \caption{\textbf{Annotation agreement on NEST.} Weighted Cohen's $\kappa$ and mean semantic similarity for GOLD–GOLD (two annotators) and GOLD–SILVER.}
  \label{fig:metrics_comparison}
\end{figure}

\paragraph{Agreement Analysis.}
We measured weighted Cohen's $\kappa \approx 0.50$ for GOLD--SILVER agreement, compared to inter-annotator $\kappa \approx 0.57$ on the GOLD set alone (Figure~\ref{fig:metrics_comparison}).
The narrow gap between these two figures suggests that the SILVER pipeline tracks human judgments at a level close to the variation between human annotators themselves.

\paragraph{Semantic Similarity Computation.}
Mean semantic similarity in Figure~\ref{fig:metrics_comparison} is computed using the same LLM-as-a-judge framework described in Appendix~\ref{sec:appendix-judge}.
For each GOLD event, we pair it with its closest SILVER event (and analogously for GOLD--GOLD pairs) and prompt GPT-5-mini~\cite{singh2025openaigpt5card} with the two event descriptions (trigger and context) to produce a similarity score between 0 and 1, using temperature 0 for reproducibility.
The reported values (0.511 for GOLD--SILVER, 0.596 for GOLD--GOLD) are macro-averaged across the five GOLD movies.
We use this LLM-based semantic similarity alongside $\kappa$ because exact string matching is too brittle for narrative event descriptions where semantically equivalent triggers may differ lexically (e.g., ``fight'' vs.\ ``attack''). Table~\ref{tab:per-field-agreement} further breaks down agreement by annotation field, covering arguments, localization, and event relations.

\begin{table}[t]
\centering
\scriptsize
\setlength{\tabcolsep}{4pt}
\begin{tabular}{lcc}
\toprule
\textbf{Field} & \textbf{Metric} & \textbf{Agreement} \\
\midrule
Arguments & PropBank role F1 & 70.86 \\
Localization & Scene-level match & 57.14 \\
Relations & Type agreement & 41.03 \\
\midrule
Overall & Weighted $\kappa$ & 
\makecell{G--G: 0.57\\G--S: 0.50} \\
\bottomrule
\end{tabular}
\caption{Per-field agreement analysis. Argument agreement measures PropBank role overlap, localization uses scene-level matching, and relation agreement measures type consistency over matched positive relation pairs. Overall agreement is reported as weighted $\kappa$ separately for GOLD--GOLD (G--G) and GOLD--SILVER (G--S) annotator pairs.}
\label{tab:per-field-agreement}
\end{table}

\paragraph{Human Effort Statistics.}
Although the GOLD subset consists of 5 movies, this corresponds to approximately 350 annotated events and 250 annotated relations.
Annotating narrative-level events in full-length movies is substantially more labor-intensive than segment-level annotation in short clips, as annotators must watch the entire film and maintain coherent tracking of characters, subplots, and temporal structure throughout.
We note that many existing benchmarks also make use of silver-labeled data, and that the scale of per-movie annotation effort in NEST is significantly higher than in benchmarks operating on short segments.
Table~\ref{tab:human-effort} summarizes the human annotation effort.
The compensation rate exceeded the applicable state minimum wage during the annotation period \cite{dol2026stateminimumwage} and was compliant with institutional policies for compensation of human subjects.

\begin{table}[ht]
\centering
\small
\begin{tabular}{lc}
\toprule
\textbf{Metric} & \textbf{Value} \\
\midrule
Number of annotators & 5 \\
Compensation rate & \$15 / hour \\
Total cost & ${\sim}$\$600 \\
Approx. hours per movie & ${\sim}$8 \\
Total annotator hours & ${\sim}$40 \\
\bottomrule
\end{tabular}
\caption{Human annotation effort for the GOLD subset. Annotators were trained via a custom platform tutorial and one practice movie before beginning production annotation.}
\label{tab:human-effort}
\end{table}

\subsection{Video Sampling and Context Handling}
\label{sec:appendix-sampling}

\paragraph{Training Token Budget.}
We fine-tune Qwen3-Omni-30B-A3B-Instruct \cite{xu2025qwen3omnitechnicalreport}, which has a native context length of 65{,}536 tokens. For training, we restrict the context budget to 32{,}768 tokens due to memory constraints.
The Qwen3-Omni vision encoder uses 16$\times$16 spatial patches with a 2$\times$2 spatial merger and a temporal patch size of 2, reducing both the spatial and temporal visual token count.
A typical movie in NEST runs approximately 2 hours (7{,}200 seconds).
At 0.1 FPS, we obtain $\sim$720 sampled frames, which are temporally grouped in pairs by the vision encoder.
The per-frame resolution is configured to keep the resulting visual tokens, text instructions, and generated outputs within the 32K training context budget.
This sampling rate prioritizes full narrative coverage over fine-grained action capture, which we find sufficient for story-level event understanding.
During training, gradient and optimizer state storage further constrains the effective batch size that fits in GPU memory.

\paragraph{Inference Token Budget.}
The nominal sampling rate does not necessarily equal the effective frame rate seen by each model because model-specific preprocessing imposes additional frame and token budgets. For Qwen3-VL and Qwen2.5-VL, a nominal 1 FPS request is uniformly subsampled to 768 frames across the full movie before inference, corresponding to approximately 0.13 FPS for an average NEST movie. The preprocessing pipeline also adapts spatial resolution to satisfy the visual-token budget, yielding approximately 106K--115K visual tokens for Qwen3-VL and 66K--74K for Qwen2.5-VL in measured examples. InternVL3.5 uses a fixed budget of 32 frames at $448\times448$ resolution. Table~\ref{tab:sampling-config} summarizes the effective sampling and preprocessing configuration for each baseline.

\begin{table*}[t]
\centering
\small
\resizebox{\textwidth}{!}{%
\begin{tabular}{lcccc}
\toprule
\textbf{Model} &
\textbf{Context} &
\textbf{Sampling} &
\textbf{Effective Frames} &
\textbf{Visual Tokens} \\
\midrule

Gemini 2.5 Pro \cite{comanici2025gemini25pushingfrontier}
& 1M & 1 FPS & $\sim$5.9K & $\sim$0.39M \\

GPT-5 \cite{singh2025openaigpt5card}
& 400K & 1 FPS & $\sim$5.9K & $\lesssim$400K \\

Qwen3-VL (8B) \cite{bai2025qwen3vltechnicalreport}
& 256K & 1 FPS & 768 & $\sim$110K \\

Qwen3-VL (30B) \cite{bai2025qwen3vltechnicalreport}
& 256K & 1 FPS & 768 & $\sim$110K \\

Qwen3-Omni (30B) \cite{xu2025qwen3omnitechnicalreport}
& 64K & 1 FPS & $\sim$450 & $\sim$64K \\

Qwen2.5-VL (7B) \cite{bai2025qwen25vltechnicalreport}
& 128K & 1 FPS & 768 & $\sim$70K \\

Qwen2.5-VL (32B) \cite{bai2025qwen25vltechnicalreport}
& 128K & 1 FPS & 768 & $\sim$70K \\

InternVL3.5 (30B) \cite{wang2025internvl35advancingopensourcemultimodal}
& 40K & 32 frames & 32 & $\sim$8.2K \\

LLaVA-Video (7B) \cite{zhang2025llavavideo}
& 32K & 64 frames & 64 & $\sim$10.8K \\

Ovis2.5 (9B) \cite{lu2025ovis25technicalreport}
& 40K & 8 frames & 8 & $\sim$6.3K \\

LongVU-LLaMA3 (3B) \cite{pmlr-v267-shen25j}
& 128K & 1 FPS & $\sim$5.9K & $\sim$8K \\

LongVU-Qwen2 (7B) \cite{pmlr-v267-shen25j}
& 32K & 1 FPS & $\sim$5.9K & $\sim$8K \\

Video-LLaMA3 (7B) \cite{zhang2025videollama3frontiermultimodal}
& 32K & 1 FPS & $\leq$180 & $\sim$16K \\

Flash-VStream-Qwen (7B) \cite{zhang2025flashvstreamefficientrealtimeunderstanding}
& 32K & 1 FPS & $\sim$5.9K & $\sim$11.5K \\

\midrule

Finetuned Qwen3-Omni \cite{xu2025qwen3omnitechnicalreport}
& 64K & 1 FPS & $\sim$450 & $\sim$64K \\

\bottomrule
\end{tabular}%
}
\caption{Video input configuration for each baseline. Sampling denotes the input sampling rate or fixed frame budget, while Effective Frames reports the resulting number of frames processed by the model. Visual Tokens reports the corresponding visual-token budget.}
\label{tab:sampling-config}
\end{table*}

\subsection{Bootstrap Confidence Intervals}
\label{sec:appendix-bootstrap}

To quantify uncertainty in selected headline benchmark results, we compute
95\% confidence intervals using movie-level bootstrap resampling.
Evaluation set sizes differ across task--model configurations because
full-length movie inference is subject to model-specific context,
frame-budget, and computational constraints; each result is therefore
computed over the subset of movies successfully processed under its
corresponding configuration. For each result, we resample these evaluation
movies with replacement, pool predictions from the sampled movies, recompute
the metric, and report the 2.5th and 97.5th percentiles.

\begin{table}[t]
\centering
\scriptsize
\setlength{\tabcolsep}{3pt}
\resizebox{\columnwidth}{!}{%
\begin{tabular}{llcc}
\toprule
\textbf{Task} & \textbf{Model} & \textbf{\#Mov.} & \textbf{Score [95\% CI]} \\
\midrule
ETD & LLaVA-Video 7B & 19 & 7.98 [6.30, 9.84] \\
EL & Qwen3-VL 30B & 25 & 5.89 [4.55, 7.41] \\
EAE & Ovis2.5 9B & 5 & 10.62 [3.47, 13.37] \\
ERE & Qwen2.5-VL 32B & 25 & 34.23 [31.89, 39.04] \\
ERE (FT) & Qwen3-Omni 30B & 5 & 49.20 [31.20, 49.26] \\
\bottomrule
\end{tabular}%
}
\caption{Movie-level bootstrap 95\% confidence intervals for selected
headline NEST results. Evaluation set sizes vary across task--model
configurations due to model-specific context, frame-budget, and computational
constraints; \#Mov.\ denotes the number of evaluation movies used as bootstrap
resampling units for each configuration.}
\label{tab:bootstrap-ci}
\end{table}

As shown in Table~\ref{tab:bootstrap-ci}, the confidence intervals reinforce
the main benchmark findings. ETD, EL, and EAE remain low even at the upper
bounds of their intervals, indicating that grounded event discovery remains
challenging. ERE achieves substantially stronger performance, while the
wider confidence interval for the fine-tuned ERE result reflects the smaller
evaluation set used for that configuration. These results support the
conclusion that relation reasoning over provided events is more tractable
than discovering and grounding narrative events in full-length video.

\subsection{LLM-as-a-Judge Evaluation Details}
\label{sec:appendix-judge}

\begin{table}[t]
\centering
\small
\setlength{\tabcolsep}{4pt}
\begin{tabular}{p{0.72\columnwidth}c}
\toprule
\textbf{Comparison} & \textbf{Agreement} \\
\midrule
GPT-5-mini \cite{singh2025openaigpt5card} vs. Gemini 2.5 Pro
\cite{comanici2025gemini25pushingfrontier} 
& 0.88 \\
\bottomrule
\end{tabular}
\caption{Judge validation on a stratified sample of approximately 200 ETD/EAE predictions.}
\label{tab:judge-validation}
\end{table}

The LLM-as-a-judge paradigm has become a widely used evaluation methodology for open-ended generation and multimodal understanding tasks.
Recent survey and empirical work has systematized this evaluation setting and shown that strong judge models can achieve substantial agreement with human judgments, while also emphasizing the importance of careful judge design and validation~\cite{gu2025surveyllmasajudge,bavaresco2025llmsinsteadhumanjudges}.
Consistent with this, GPT-5-mini and Gemini 2.5 Pro achieve 0.88 agreement on our validation sample, as shown in Table~\ref{tab:judge-validation}.
In the video understanding domain, LLM-based evaluators have become common for open-ended question answering and reasoning benchmarks where exact string matching is too brittle to capture semantic equivalence~\cite{Maaz2023VideoChatGPT,nagrani2025minervaevaluatingcomplexvideo,shaar2026movierecapsqamultimodalopenendedvideo}.
We follow this established practice for NEST, where strict string matching is too brittle for open-ended tasks such as event argument extraction, in which semantically equivalent mentions may differ lexically (e.g., ``the detective'' vs.\ ``Officer Miller'').

\paragraph{Judge Model and Settings.}
We use GPT-5-mini~\cite{singh2025openaigpt5card} as the judge.
\footnote{The judge model is not among the evaluated baselines, except that GPT-5 is evaluated separately as a baseline from the same model family.}
Decoding is performed with temperature $0$ (greedy) to maximize reproducibility.
The judge receives the ground-truth annotation and the model prediction, and returns a binary verdict (correct/incorrect) along with a confidence score.

\paragraph{Judge Prompts.}
The full judge prompts for each task are provided below and 
each prompt follows a general template consisting of: (1) the task definition, (2) evaluation rules specifying what constitutes a correct prediction, (3) output format rules, and (4) the ground-truth and model prediction as inputs.
The task-specific definitions and rules are identical to those used during model evaluation, ensuring consistency between how models are prompted and how their outputs are judged. The generic judge template and the task-specific ETD and EAE judge prompts are provided in Figures~\ref{fig:judge-template}, \ref{fig:etd-judge-template}, and~\ref{fig:eae-judge-template}.

\subsection{Qualitative Examples and Error Analysis}
\label{sec:appendix-qualitative}

\begin{figure*}[t]
    \centering
    \includegraphics[width=1\linewidth]{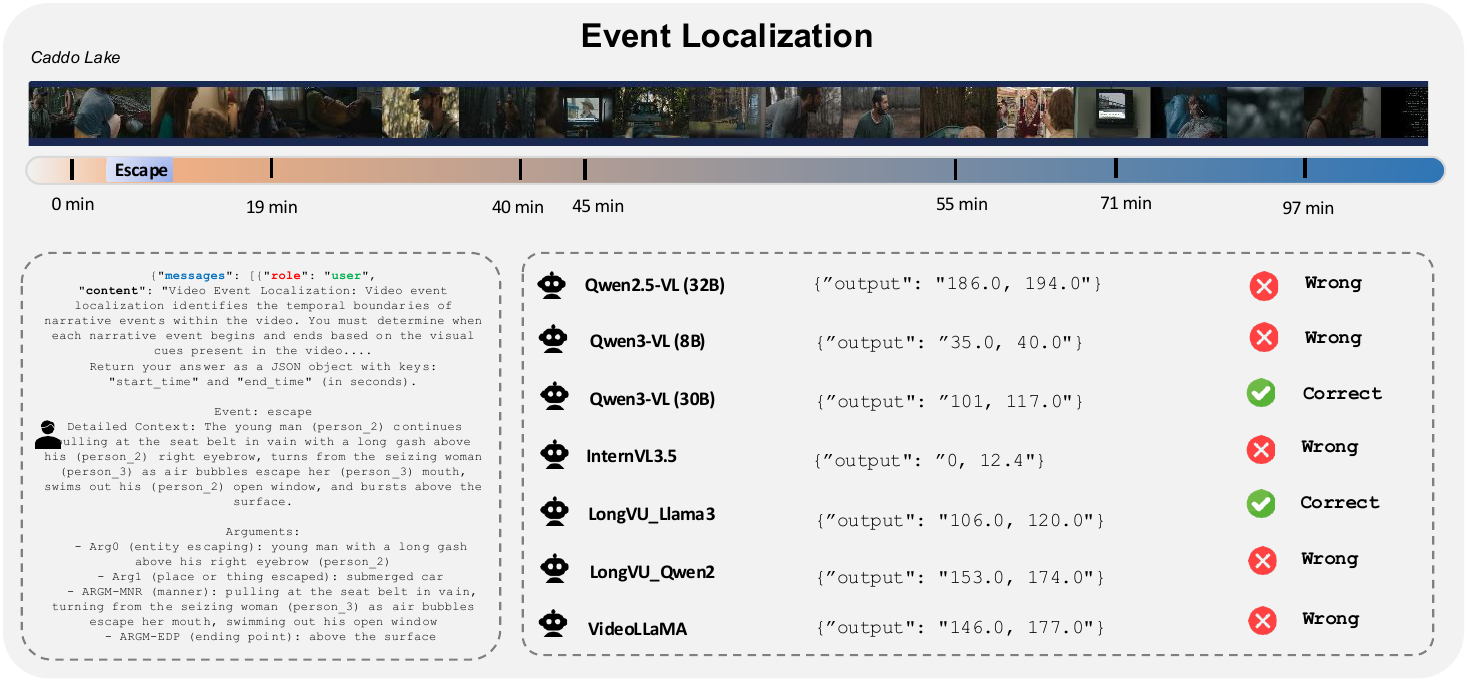}
    \caption{Event Localization example from \textit{Caddo Lake} (97 min). The ``escape'' event occurs at approximately 101--120 seconds into the movie. Given the full movie and the event description with structured arguments (left panel), seven models predict temporal boundaries. Only Qwen3-VL (30B), predicting 101--117 seconds, and LongVU-LLaMA3, predicting 106--120 seconds, correctly localize the event. Qwen3-VL (8B) and InternVL3.5 predict timestamps that are too early, while Qwen2.5-VL (32B), LongVU-Qwen2, and VideoLLaMA predict later intervals within the first few minutes of the movie. The example illustrates that even when predictions fall near the correct portion of a full-length movie, models often fail to identify the correct event boundaries.}
    \label{fig:el-example}
\end{figure*}

\begin{figure*}[t]
    \centering
    \includegraphics[width=1\linewidth]{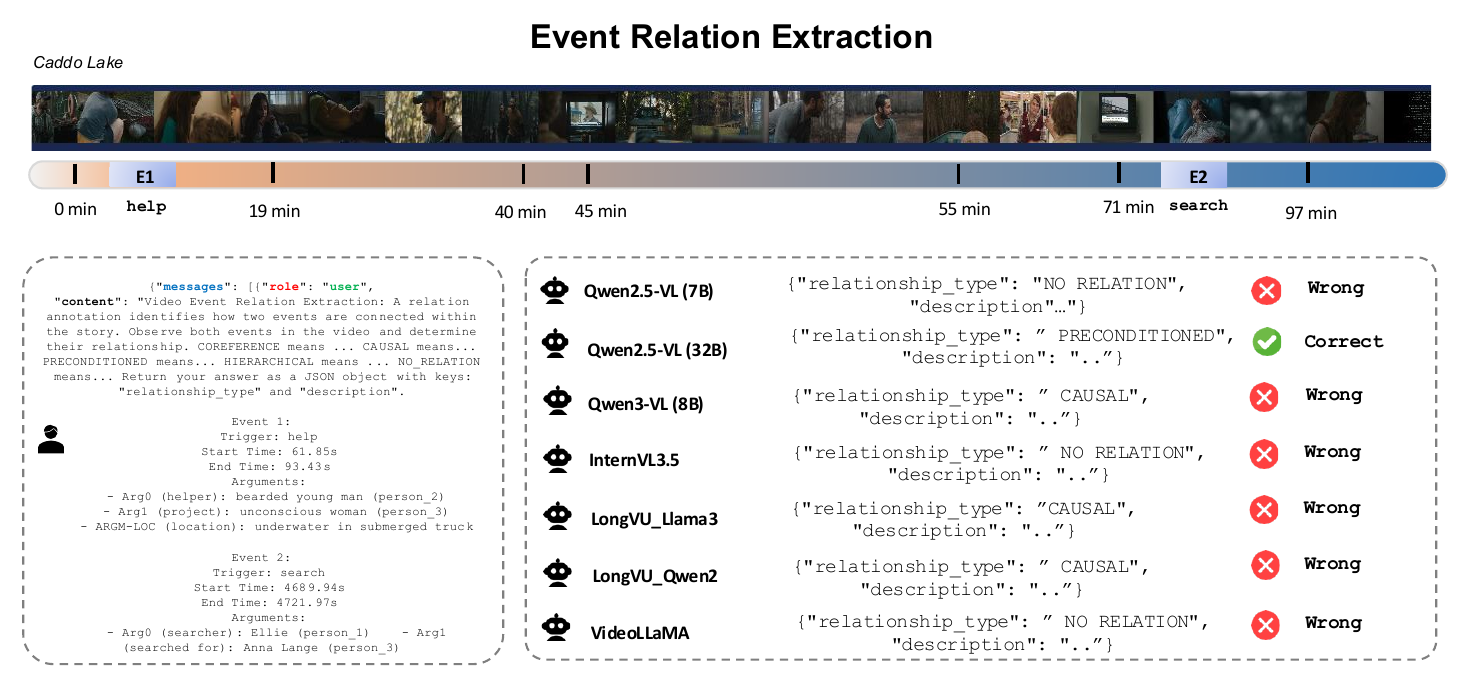}
    \caption{Event Relation Extraction example from \textit{Caddo Lake} (97 min). Two events are separated by approximately 76 minutes: a ``help'' event (E1, near 2 min) and a ``search'' event (E2, near 78 min). The ground-truth relation is PRECONDITIONED, as the earlier helping event creates conditions that enable the later search. Only Qwen2.5-VL (32B) correctly identifies this relation. Three models (Qwen3-VL, LongVU-LLaMA3, LongVU-Qwen2) predict CAUSAL, confusing enablement with direct causation. Three others (Qwen2.5-VL 7B, InternVL3.5, VideoLLaMA) predict NO\_RELATION entirely, failing to connect events separated by long temporal gaps. This illustrates the challenge of reasoning over distant event pairs where the narrative link requires understanding how an earlier event sets up conditions for a later one.}
    \label{fig:ere-preconditioned-example}
\end{figure*}

\begin{figure*}[t]
    \centering
    \includegraphics[width=1\linewidth]{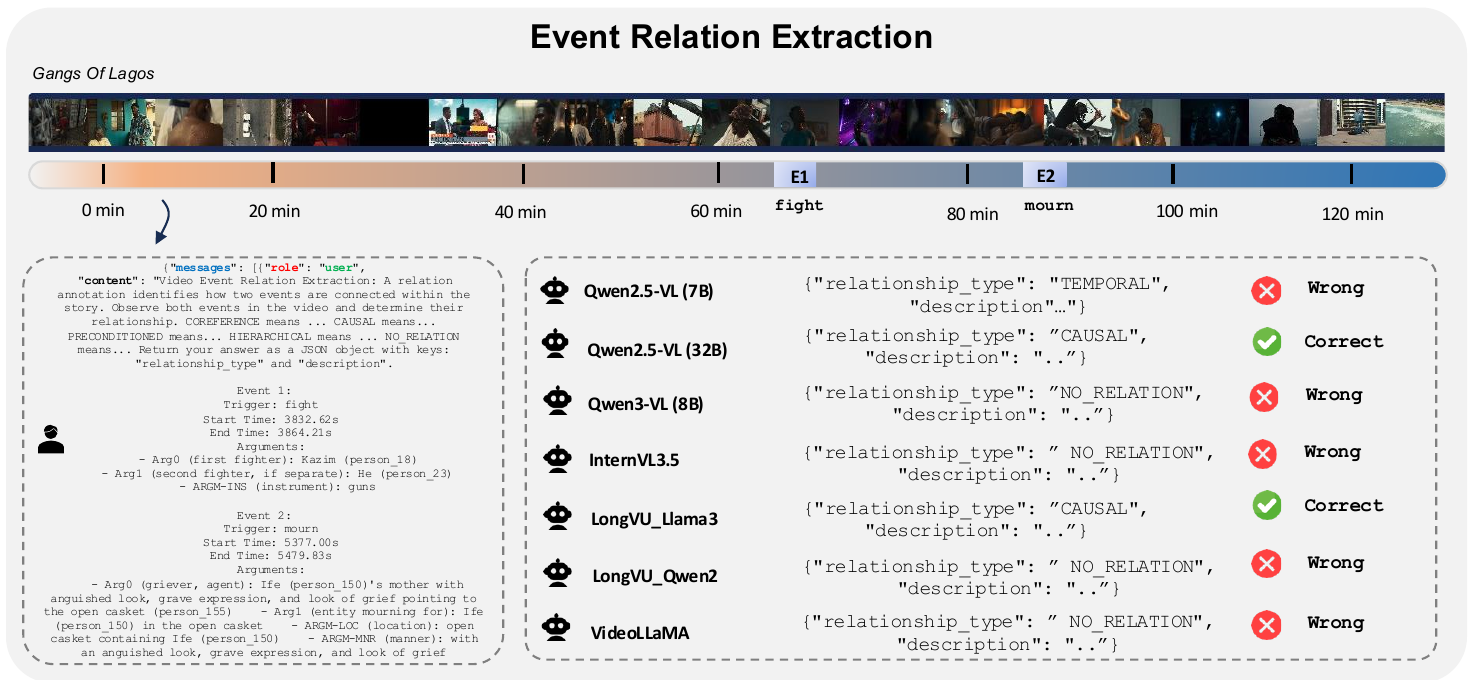}
    \caption{Event Relation Extraction example from \textit{Gangs of Lagos} (120 min). A ``fight'' event (E1, near 64 min) is followed by a ``mourn'' event (E2, near 90 min), with the ground-truth relation being CAUSAL. Only Qwen2.5-VL (32B) and LongVU-LLaMA3 correctly identify the causal link. Qwen2.5-VL (7B) predicts TEMPORAL, recognizing the sequential ordering but missing the causal dependency. The remaining four models predict NO\_RELATION, failing to connect the two events despite their relative temporal proximity ($\sim$26 minutes apart). This example highlights that even when events are not separated by extreme temporal gaps, most models still struggle to infer narrative causality from video content.}
    \label{fig:ere-causal-example}
\end{figure*}

\begin{figure*}[t]
    \centering
    \includegraphics[width=1\linewidth]{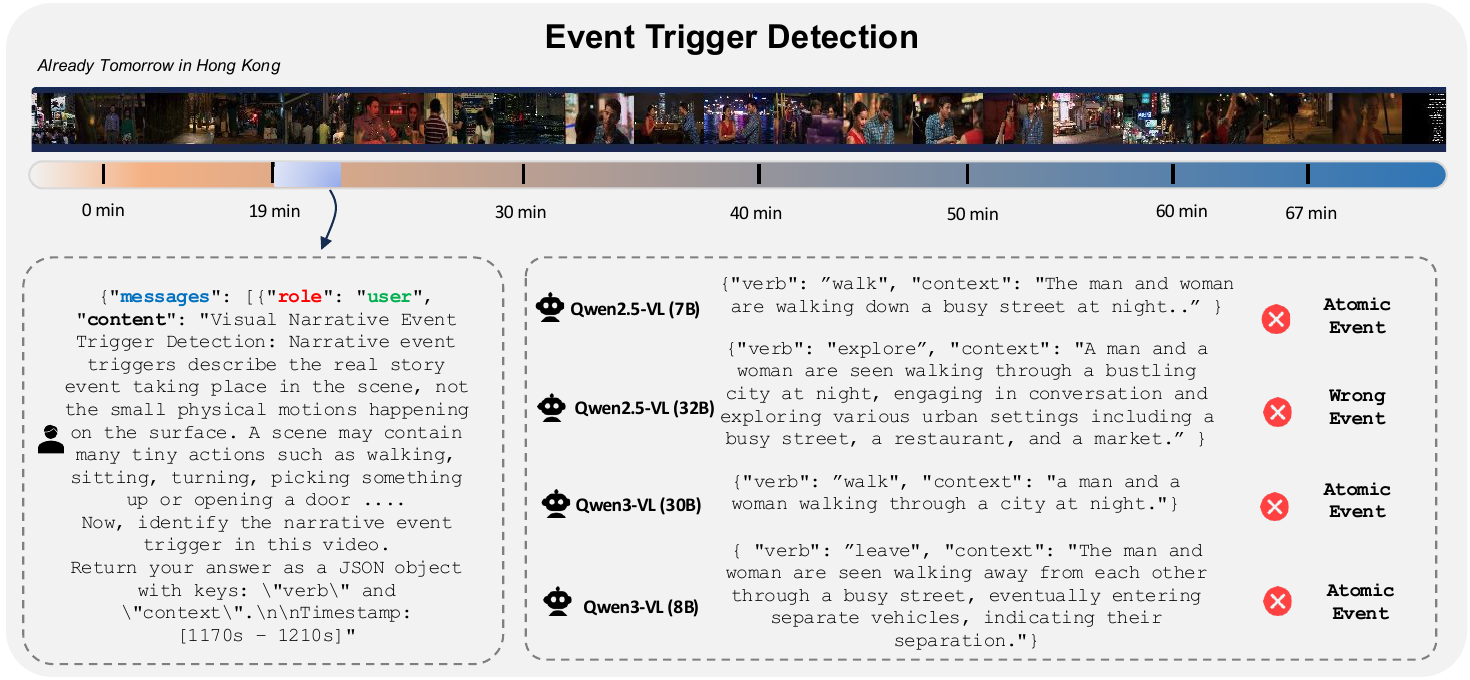}
    \caption{Event Trigger Detection example from \textit{Already Tomorrow in Hong Kong} (67 min). Given the scene between 1170--1210 seconds, all four models fail to identify the correct narrative event. Three out of four models (Qwen2.5-VL 7B, Qwen3-VL 30B, and Qwen3-VL 8B) predict atomic verbs (``walk'' or ``leave''), describing surface-level physical motions rather than the narrative-level event taking place. Qwen2.5-VL (32B) predicts ``explore'', which is closer to a narrative description but still misses the intended event. Despite the prompt explicitly instructing models to identify the real story event rather than small physical motions, all models default to describing what is visually immediate. This example highlights the atomic verb default as the dominant ETD failure mode and suggests that current models lack the narrative reasoning needed to abstract from visual observations to story-level meaning.}
    \label{fig:etd-example-2}
\end{figure*}

We present representative examples of NEST annotations alongside model predictions to illustrate common failure modes.
Figures~\ref{fig:el-example}--\ref{fig:etd-example-2} show concrete examples for each task.
To quantify these patterns, we analyze errors from five representative zero-shot models. For ETD, EAE, and EL, we analyze all incorrect predictions available for each model. For ERE, we analyze how often ground-truth related event pairs are incorrectly assigned NO\_RELATION. Per-model sample sizes are reported in Table~\ref{tab:error-taxonomy}.

\paragraph{Event Trigger Detection Errors.}
A frequent failure mode in ETD is that models default to surface-level atomic verbs (e.g., ``walk,'' ``look,'' ``sit'') rather than identifying the intended narrative-level predicate (e.g., ``confront,'' ``betray,'' ``reconcile'').
Across the five analyzed models, atomic-verb defaults account for 20.8\% of errors on average, while wrong-event predictions account for 76.4\%.
The atomic verb default occurs especially when long-range context is required to understand the narrative significance of a scene.
For example, a scene showing two characters meeting in a park may be annotated as ``reconcile'' (requiring understanding of their earlier conflict), but models predict ``meet'' or ``talk'' based on immediate visual cues alone.
The dominance of wrong-event errors (76.4\% on average) suggests that models frequently fail to identify the correct narrative event at all, rather than merely selecting the wrong level of abstraction.

\paragraph{Event Argument Extraction Errors.}
In EAE, entity confusion dominates across the five analyzed models, accounting for 97.6\% of errors on average, while missing arguments account for only 2.4\%.
Models frequently fail to maintain consistent entity references across distant scenes.
A character introduced by name in an early scene may be referred to by description (``the detective'') or pronoun (``he'') in later predictions, and models often assign arguments to the wrong character when multiple characters share similar visual appearances.
The overwhelming prevalence of entity confusion over missing arguments indicates that models attempt to fill argument slots but lack the long-range entity tracking needed to do so correctly.

\begin{table}[ht]
\centering
\small
\begin{tabular}{lc}
\toprule
\textbf{Task} & \textbf{Score (\%)} \\
\midrule
ETD Accuracy             & 12.50 \\
EAE Accuracy             & 25.00 \\
EL Accuracy              & 2.50 \\
\midrule
ERE Precision            & 40.95 \\
ERE Recall               & 44.64 \\
ERE F1 (macro)           & 40.80 \\
ERE Accuracy             & 55.56 \\
\midrule
Overall Accuracy (macro) & 23.89 \\
\bottomrule
\end{tabular}
\caption{GPT-5 performance across all four NEST tasks, evaluated on the same 10-video subset used for Gemini in Table~\ref{tab:event-relation}. ETD and EAE use LLM-as-a-judge accuracy, EL uses scene-overlap accuracy, and ERE reports Precision and Recall at the prediction level along with macro F1 across the six relation types. Overall Accuracy is macro-averaged across the four task accuracies.}
\label{tab:gpt5-nest-sampled}
\end{table}

\paragraph{Event Localization Errors.}
For EL, nearly all errors across the five analyzed models are wrong-scene predictions, averaging 98.4\%, while overly broad time spans account for only 1.6\%.
Models tend to localize events to the most visually salient scene (which may not be the correct one) rather than hedging with broad spans.
Flashback sequences are particularly challenging: models consistently assign flashback events to the narrative present rather than recognizing the temporal displacement.
The near-absence of overly broad spans suggests that models are confident but wrong, committing to specific (incorrect) temporal locations rather than expressing uncertainty through wider predictions.

\paragraph{Event Relation Extraction Errors.}
As shown in Table~\ref{tab:error-taxonomy}, ERE errors are often dominated by false-negative NO\_RELATION predictions, although the extent varies substantially by model. InternVL3.5 incorrectly predicts NO\_RELATION for 78.3\% of ground-truth related event pairs. Similar behavior is observed for Qwen2.5-VL (32B), Qwen3-VL (8B), and Qwen3-VL (30B), with rates of 73.3\%, 83.7\%, and 92.2\%, respectively, while Qwen2.5-VL (7B) is a notable exception at 21.2\%. This pattern suggests that most models are overly conservative in identifying narrative connections and often default to predicting no relation between events.

\begin{table}[t]
\centering
\scriptsize
\begin{tabular}{lcc}
\toprule
\textbf{Model} & \textbf{\#Params} & \textbf{F1 (\%)} \\
\midrule
Flash-VStream-Qwen & 7B & 0.00 \\
InternVL3.5 & 30B & 0.00 \\
LLaVA-Video & 7B & 0.00 \\
Qwen2.5-VL & 32B & 28.57 \\
Qwen2.5-VL & 7B & 13.04 \\
Qwen3-VL & 30B & 0.00 \\
Qwen3-VL & 8B & 0.00 \\
\bottomrule
\end{tabular}
\caption{Performance on the NEST flashback subset. These relations capture non-linear temporal structure and require models to reason beyond linear timelines. }

\label{tab:f1_flashback_subset}
\vspace{-1em}
\end{table}

\begin{table}[ht]
\centering
\small
\begin{tabular}{llc}
\toprule
\textbf{Task} & \textbf{Error Type} & \textbf{\%} \\
\midrule
ETD & Atomic event & 20.8 \\
ETD & Wrong event entirely & 76.4 \\
\midrule
EAE & Entity confusion & 97.6 \\
EAE & Missing argument & 2.4 \\
\midrule
EL & Wrong scene & 98.4 \\
EL & Overly broad span & 1.6 \\
\bottomrule
\end{tabular}
\caption{Average error taxonomy across the five models analyzed in Table~\ref{tab:error-taxonomy}. Values are macro-averaged across models for each error type.}
\label{tab:error-taxonomy-models}
\end{table}

\begin{table*}[t]
\centering
\small
\resizebox{\textwidth}{!}{%
\begin{tabular}{l|ccc|ccc|ccc|cc}
\toprule
\multirow{2}{*}{\textbf{Model}} &
\multicolumn{3}{c|}{\textbf{ETD}} &
\multicolumn{3}{c|}{\textbf{EAE}} &
\multicolumn{3}{c|}{\textbf{EL}} &
\multicolumn{2}{c}{\textbf{ERE}} \\
& $N$ & Atomic Verb & Wrong Event
& $N$ & Entity Conf. & Missing Arg.
& $N$ & Wrong Scene & Broad Span
& $N$ & $\rightarrow$NO\_REL \\
\midrule
InternVL3.5 \cite{wang2025internvl35advancingopensourcemultimodal}
& 1812 & 22.0 & 78.0
& 1779 & 90.0 & 10.0
& 1857 & 98.6 & 1.4
& 1604 & 78.3 \\

Qwen2.5-VL (7B) \cite{bai2025qwen25vltechnicalreport}
& 1584 & 42.0 & 58.0
& 1779 & 100.0 & --
& 1782 & 99.1 & 0.9
& 1470 & 21.2 \\

Qwen2.5-VL (32B) \cite{bai2025qwen25vltechnicalreport}
& 1833 & 6.0 & 94.0
& 1779 & 100.0 & --
& 1863 & 97.4 & 2.6
& 1541 & 73.3 \\

Qwen3-VL (8B) \cite{bai2025qwen3vltechnicalreport}
& 1814 & 16.0 & 82.0
& 1779 & 98.0 & 2.0
& 1851 & 99.4 & 0.6
& 1604 & 83.7 \\

Qwen3-VL (30B) \cite{bai2025qwen3vltechnicalreport}
& 1805 & 18.0 & 70.0
& 1779 & 100.0 & --
& 1761 & 97.6 & 2.4
& 1604 & 92.2 \\
\bottomrule
\end{tabular}%
}
\caption{Error taxonomy across models. For ETD, EAE, and EL, $N$ denotes the number of incorrect predictions analyzed, and values report the percentage exhibiting each listed error type. For ERE, $N$ denotes the number of ground-truth related event pairs, and the NO\_REL column reports the percentage incorrectly predicted as NO\_RELATION. Wrong-scene predictions consistently dominate EL errors and entity confusion dominates EAE errors. For four of the five analyzed models, more than half of ground-truth related event pairs are incorrectly classified as NO\_RELATION, while Qwen2.5-VL (7B) is a notable exception.}
\label{tab:error-taxonomy}
\end{table*}


\begin{figure}
    \centering
    \includegraphics[width=1\linewidth]{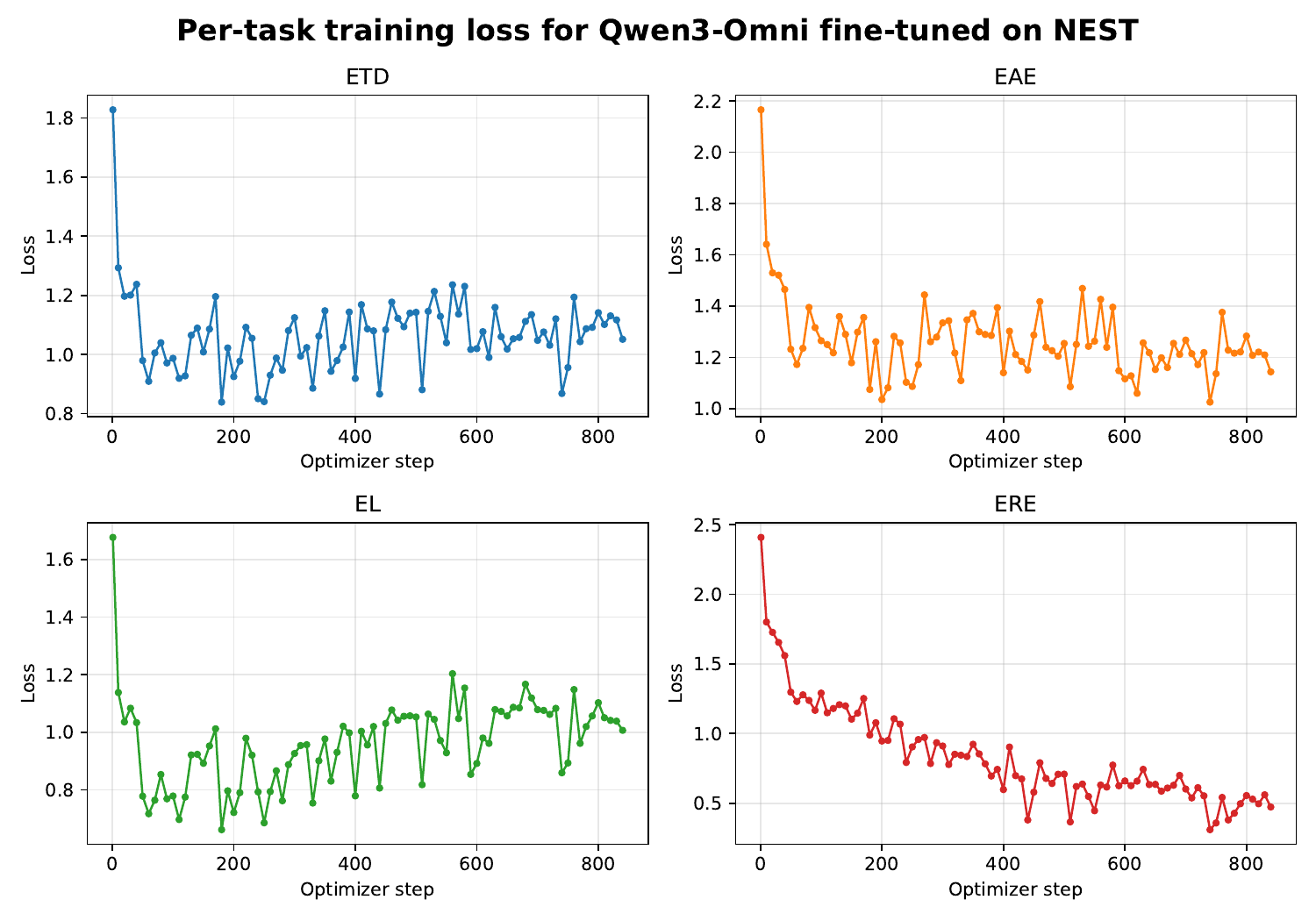}
    \caption{Task-specific training loss curves for Qwen3-Omni fine-tuned on NEST. ERE loss decreases steadily, consistent with the substantial improvement in ERE F1 after fine-tuning. ETD and EL losses plateau early, indicating that LoRA fine-tuning does not acquire the long-range narrative event induction and temporal grounding capabilities these tasks require.}
    \label{fig:task-loss}
\end{figure}

\subsection{Directions for Future Work}
\label{sec:appendix-future}

\paragraph{Human Agreement Varies Across Annotation Components.}
Human annotations on the GOLD set achieve moderate overall agreement, with weighted Cohen's $\kappa \approx 0.57$, while consistency varies across annotation components. Agreement is higher for argument roles (70.86 F1) and scene-level localization (57.14\%) than for relation types (41.03\%), reflecting the greater ambiguity involved in interpreting narrative relationships between events (Table~\ref{tab:per-field-agreement}). These results characterize the consistency of the annotation process rather than human performance on the four benchmark tasks and therefore should not be interpreted as a human performance ceiling.

\paragraph{SFT Alone Is Insufficient.}
Fine-tuning on NEST improves ERE (19.13\% $\rightarrow$ 49.20\% F1) but yields only marginal gains on ETD (3.20\% $\rightarrow$ 6.09\%) and EL (0.44\% $\rightarrow$ 0.45\%). As shown in Figure~\ref{fig:task-loss}, ERE loss continues to decrease during training, while ETD and EL losses plateau much earlier, providing further evidence that supervised fine-tuning alone is insufficient for grounded narrative event discovery and temporal localization.

\paragraph{Research Directions.}
Closing the gap between human and model performance on NEST likely requires advances in several areas: (1) narrative abstraction mechanisms that lift low-level observations into story-level events via explicit reasoning or chain-of-thought over prior context; (2) long-range entity tracking that combines visual re-identification with dialogue-based coreference across full movies; (3) non-linear temporal reasoning to handle flashbacks and parallel storylines, where current models achieve near-zero F1 (Table~\ref{tab:f1_flashback_subset});  (4) retrieval-augmented grounding that selectively attends to relevant past scenes without requiring prohibitively long context windows; and (5) hybrid symbolic-neural event structures that combine neural perception with structured representations such as event graphs or temporal logic.

\subsection{Annotating Platform Interface}
\label{sec:appendix-platform-ui}

We built a custom web-based annotation platform for NEST.
The interface allows annotators to: (1) view individual scenes from each movie in an embedded video player, (2) select event triggers from the PropBank verb vocabulary with sense disambiguation, (3) fill argument roles (ARG0, ARG1, ARGM-LOC, ARGM-TMP, etc.) with free-form mentions tagged by modality and grounding scope, and (4) annotate pairwise relations between events with supporting evidence.
Annotators were provided with a written tutorial describing the annotation guidelines, event ontology, and relation definitions, followed by a supervised practice session on one complete movie before beginning production annotation.
The platform enforces PropBank constraints at input time, preventing annotators from entering free-form triggers or unsupported relation types.

\subsection{Annotator Instructions and Prompt Templates}
\label{sec:appendix-platform-prompts}

These are the instructions provided to annotators for each annotation task.
The instructions distinguish between visual, audio, and dialogue event triggers, and provide concrete examples of correct and incorrect annotations.
Each instruction set emphasizes the distinction between atomic physical actions and narrative-level events. The complete annotator instructions for visual, audio, and dialogue event trigger detection, event argument extraction, and event relation extraction are provided in Figures~\ref{fig:visual-etd-instructions}, \ref{fig:audio-etd-instructions}, \ref{fig:dialogue-etd-instructions}, \ref{fig:eae-instructions-part1}, \ref{fig:eae-instructions-part2}, and~\ref{fig:ere-instructions}.

\subsection{Task Definitions}
\label{sec:appendix-task-definitions}

\begin{figure*}[t]
    \centering
    \captionsetup[subfigure]{labelformat=empty, font=scriptsize, skip=1pt}

    \begin{subfigure}[t]{0.98\textwidth}
        \centering
        \includegraphics[width=\linewidth]{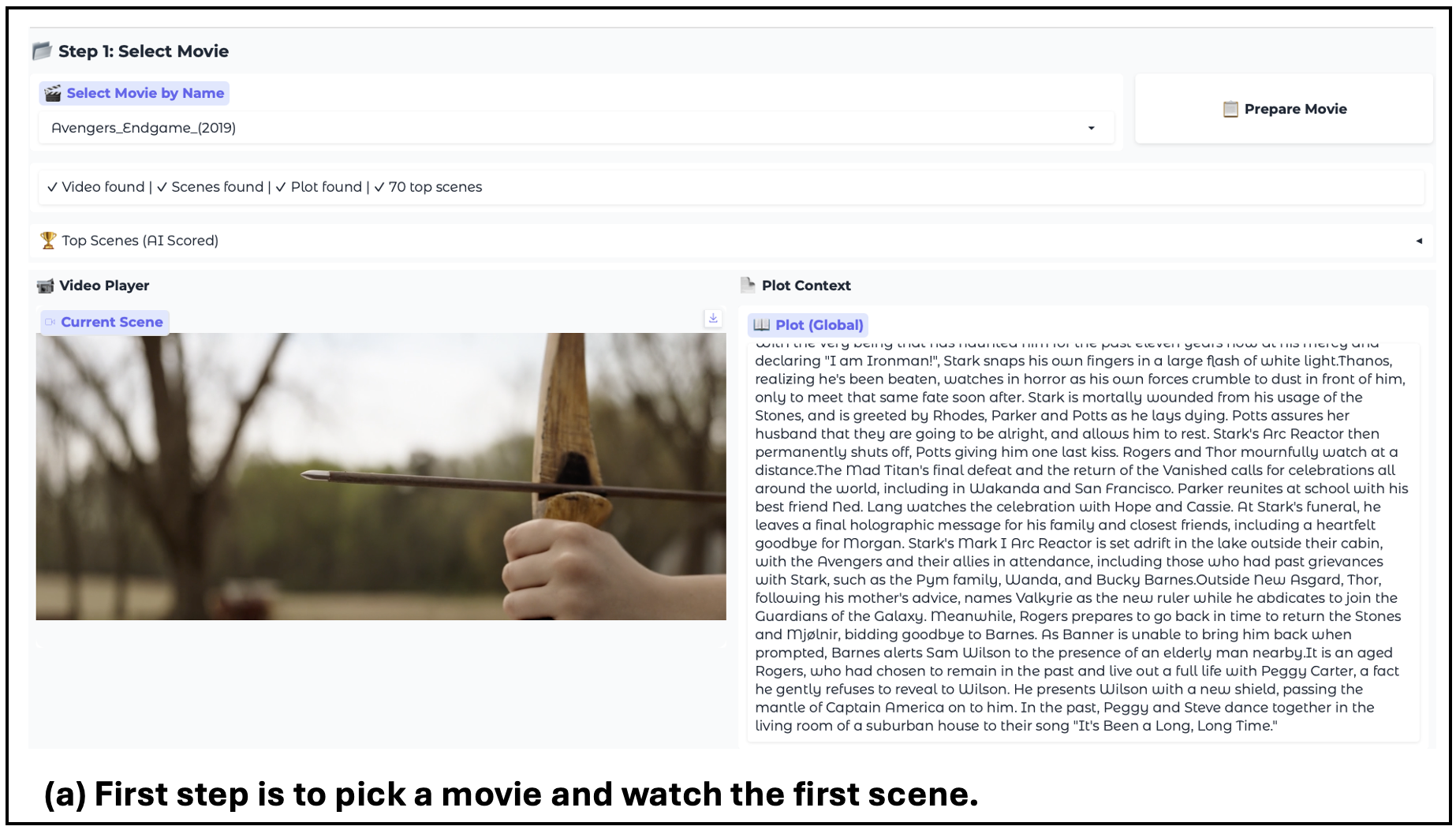}
        \caption{The annotator initially selects and loads a movie. The workspace presents the current scene in a video player, allowing the annotator to ground decisions in both local visual evidence and broader narrative context while stepping through scenes.}
    \end{subfigure}\vspace{0.35em}

    \begin{subfigure}[t]{0.98\textwidth}
        \centering
        \includegraphics[width=\linewidth]{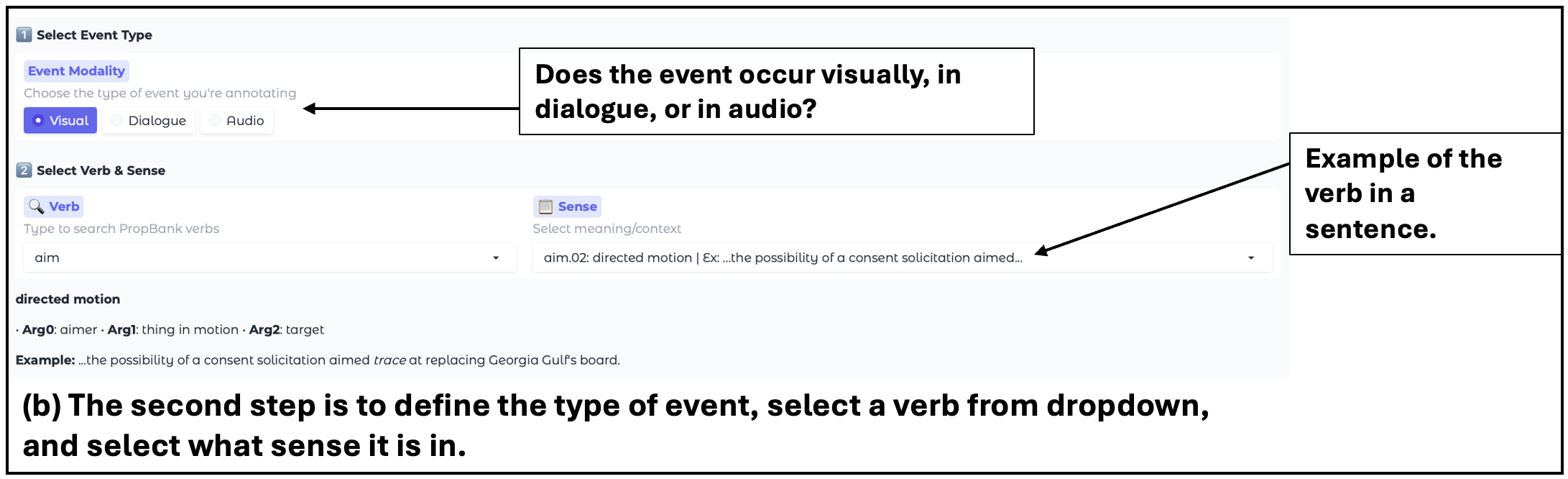}
        \caption{The annotator first selects the evidence channel (visual, dialogue, or audio), then chooses a PropBank predicate and sense. The roleset description and example usage are shown to disambiguate meaning.}
    \end{subfigure}

    \caption{NEST annotation interface for movie selection and narrative event trigger annotation. Annotators inspect individual scenes in the context of the full movie, select the evidence modality, and assign a PropBank predicate and sense to each narrative event.}
    \label{fig:nest-ui-1}
\end{figure*}

\begin{figure*}[t]
    \centering
    \captionsetup[subfigure]{font=scriptsize, skip=1pt, labelformat=empty}

    \begin{subfigure}[t]{0.98\textwidth}
        \centering
        \includegraphics[width=\linewidth]{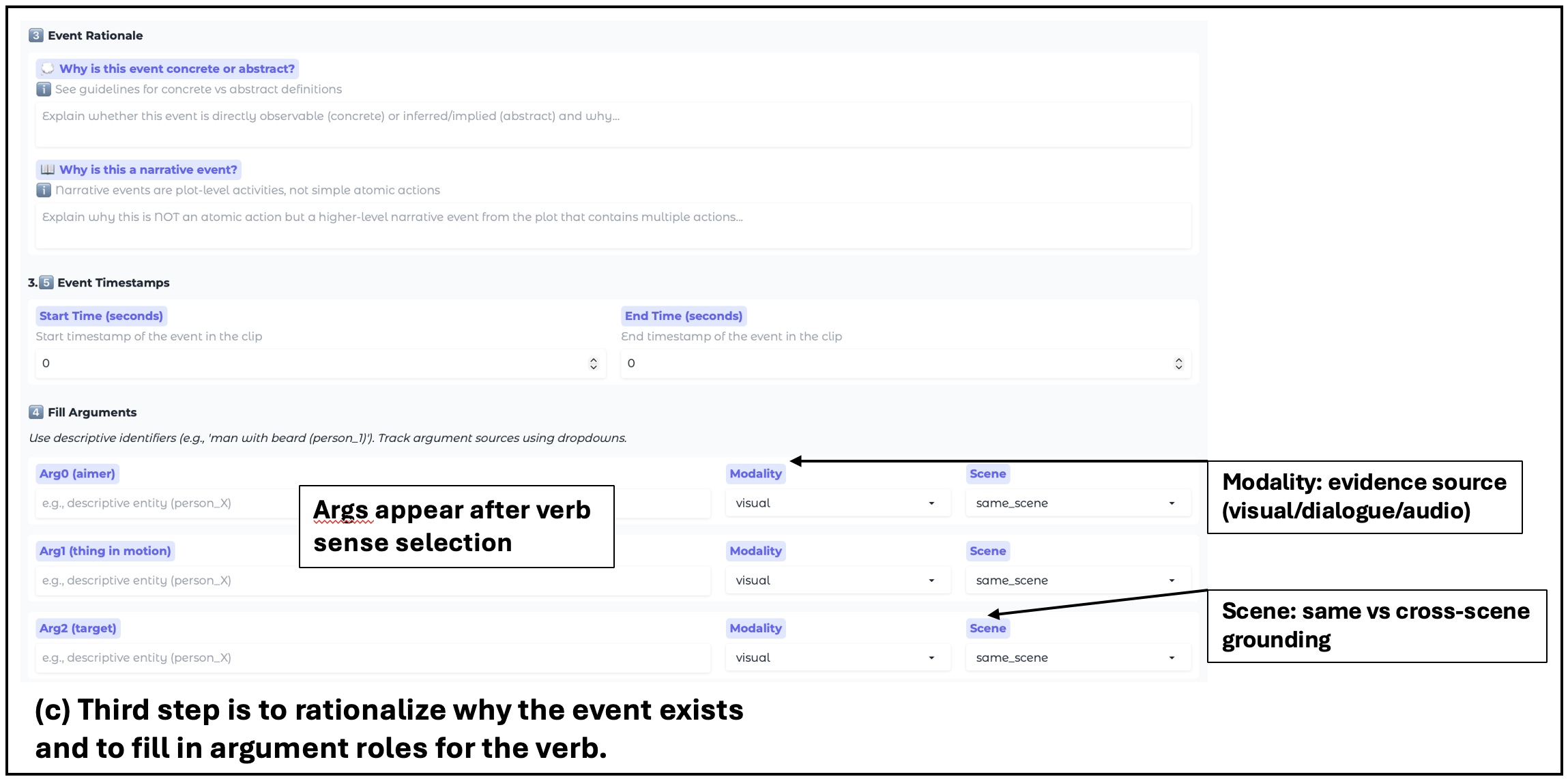}
        \caption{The annotator marks the event's temporal span and fills PropBank argument slots with free-form mentions, tagging each by modality and grounding scope. They then add the event to the scene and view all created events.}
    \end{subfigure}

    \begin{subfigure}[t]{0.98\textwidth}
        \centering
        \includegraphics[width=\linewidth]{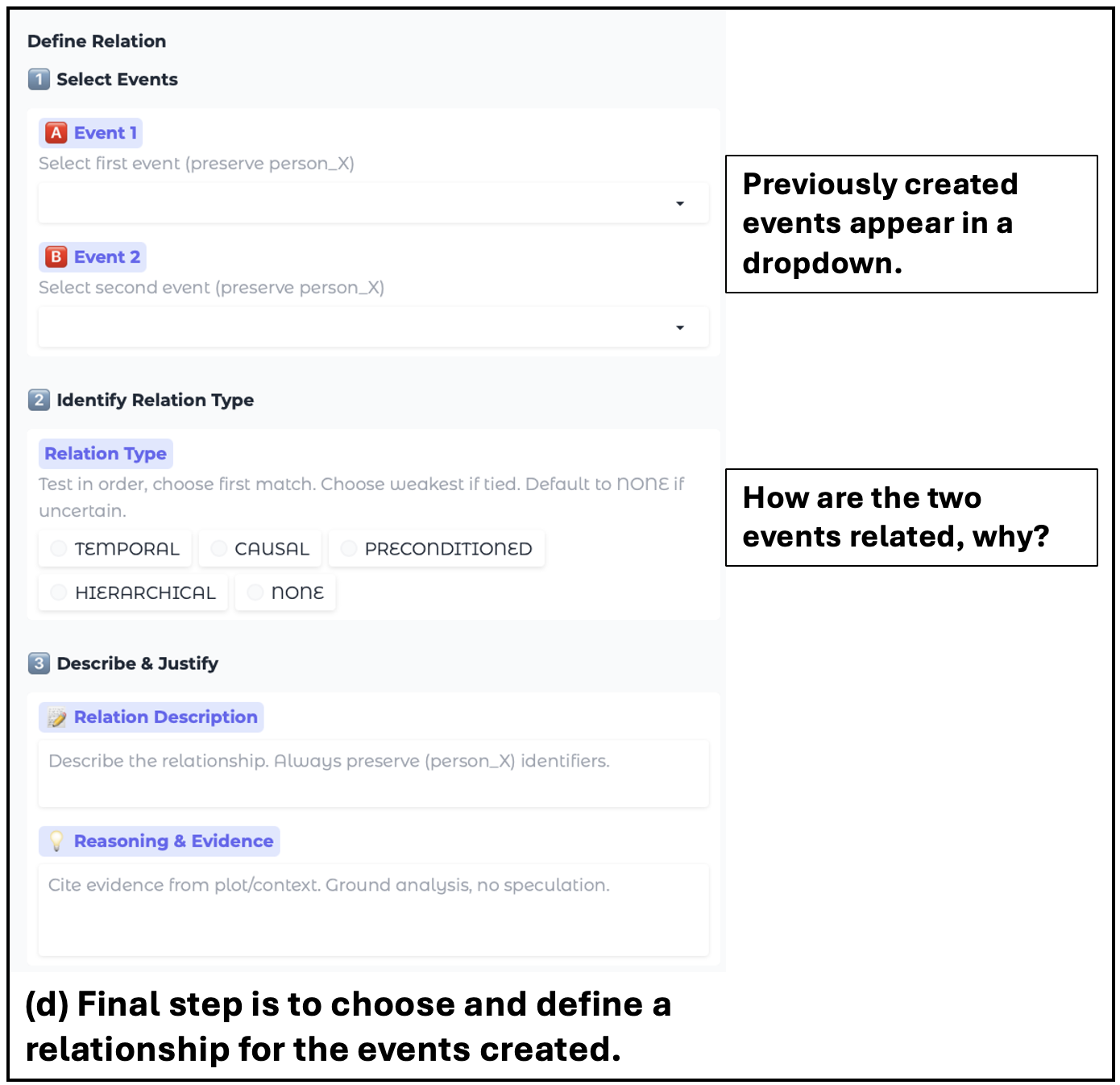}
        \caption{The annotator selects two events from the scene-level event inventory, assigns a relation type such as temporal, causal, preconditioned, hierarchical, or no relation, and records a short relation description with supporting evidence.}
    \end{subfigure}

    \caption{NEST annotation interface for event grounding, argument extraction, and event relation annotation.}
    \label{fig:nest-ui-2}
    \end{figure*}

Below we provide the JSON-formatted prompt templates for each evaluation task. The templates specify task instructions derived from our annotation guidelines and define the expected response format. Figures~\ref{fig:nest-ui-1} and~\ref{fig:nest-ui-2} show how the corresponding event triggers, arguments, temporal grounding, and relations are defined and collected in the NEST annotation interface. All evaluated models receive identical prompts to ensure fair comparison. The exact prompt templates used for ETD, EAE, ERE, and EL are provided in Figures~\ref{fig:etd-prompt-template}, \ref{fig:eae-prompt-template}, \ref{fig:ere-prompt-template}, and~\ref{fig:el-prompt-template}.


\begin{figure*}[t]
\centering
\begin{tcolorbox}[colback=blackcolor,colframe=black!75!white, title=Visual Narrative-Level Event Trigger Detection]
 
Narrative event triggers describe the real story event taking place in the scene, not the small physical motions happening on the surface. A scene may contain many tiny actions such as walking, sitting, turning, picking something up or opening a door. These movements show how someone moves, but they do not explain what the moment means. A narrative event is the larger action that these motions add up to. Someone standing up, taking a suitcase and walking out is not performing three events and together these actions mean the person leaves. Someone stepping forward, pointing and blocking another person's path adds up to confront. Someone gathering belongings and closing a door behind them adds up to leave. A single narrative event can contain several visible motions, yet the motions themselves are not the event. Break up may involve looking away, raising a voice and walking out, but the event is the end of the relationship. To find the correct narrative event, first observe the physical actions without interpreting them. Then consider what these motions together accomplish in the story. Their combined meaning may show that someone leaves, helps another person, refuses something or apologizes. Once the story-level action is clear, open the trigger dropdown on the platform and select the PropBank verb that best expresses this meaning. You do not type anything manually. You simply choose the verb that captures what the moment actually represents, and then select the correct sense that appears afterward.
The final step is to fill in the rationale fields. You must always choose a concrete visual narrative event rather than an abstract internal state such as think, realize or feel. A concrete event is something that a viewer can see happening directly in the video. When selecting the trigger, the annotator should always pause and ask whether the event is visually observable as an action. Narrative events such as kill, leave, save, confront, break up or comfort are all valid because the viewer can see the actions that form these events. For example, kill is visible when one person physically attacks another and the plot confirms the consequence. Break up is visible when someone raises their voice, withdraws, turns away and walks out, and the plot indicates the end of the relationship. Leave is visible when the character gathers belongings, exits and closes the door. These are all narrative events grounded in what the viewer sees.
You will also explain why the chosen verb is a narrative event instead of a small atomic action. This explanation should describe how the visible actions combine into one meaningful story moment and how the plot supports it. You should list the small actions that make up the narrative event you choose. A narrative event that you chose  must be mentioned in the plot either directly or indirectly. Your reasoning should point to how the plot supports the event you selected. 
Example 1: A woman packs clothing into a suitcase, pauses at the doorway, looks back once, then walks out and closes the door behind her.
Correct trigger: leave
Why: The small actions such as packing, walking and closing the door combine into the narrative meaning that she leaves the place. The situation changes because she is no longer there.
Incorrect trigger chosen: walk
Why this is wrong: Walk describes only a physical motion. It ignores the combined meaning that she is leaving the space.
Example 2: A man kneels beside an injured friend, lifts them up carefully, supports their weight and carries them toward safety.
Correct trigger: help
Why: The combined actions show one person assisting another in a meaningful way. The situation changes because the injured person receives help.
Incorrect trigger chosen: lift
Why this is wrong: Lift is one tiny part of the action. It does not capture the meaningful event of helping the injured friend.
Example 3: Two siblings sit quietly until one suddenly moves closer, places a hand on the other's shoulder and gently guides them into an embrace as the second sibling begins to cry.
Correct trigger: comfort
Why: The combined actions show one person offering emotional support. The important moment is the comforting interaction, not the individual gestures.
Incorrect trigger chosen: touch
Why this is wrong: Touch describes only one small motion. It does not reflect the meaningful event where one sibling comforts the other.
\end{tcolorbox}
\caption{Annotator instructions for visual narrative-level event trigger detection.}
\label{fig:visual-etd-instructions}
\end{figure*}

\begin{figure*}[t]
\centering
\begin{tcolorbox}[colback=blackcolor,colframe=black!75!white, title=Audio Narrative-Level Event Trigger Detection]
 
Audio event triggers label clear and identifiable actions that can be heard. An audio segment may contain many soft or vague noises such as rustling, quiet breathing, fabric movement or a steady background hum. These sounds do not point to a specific action and should not be labeled. A valid audio event is something the listener can confidently recognize as an action, such as knocking, crying, shouting, laughing or something breaking. The trigger must always be selected from the PropBank list shown in the platform. You do not write your own verb. You choose the verb from the dropdown that matches the action you hear. After selecting the trigger, the platform will display the possible senses for that verb and you must choose the correct one.
A single audio event may contain several distinct sounds, but these sounds work together to represent one action. A break event may include a sharp crash followed by objects scattering. A knock event may include a sequence of firm hits. A cry event may include whimpers that grow into sobs. To choose the correct event, first listen to the sounds without interpreting anything extra. Then decide what action the combined sounds clearly represent. Once the action is clear, open the trigger dropdown and select the PropBank verb that matches it.
For the rationale fields, you must confirm that the event is something that can be heard directly. Internal or mental states cannot be used as audio triggers. Actions like knock, break, cry, shout and laugh work because the sound itself reveals them. You must also explain why the chosen verb fits the audio. 
Example 1: You hear someone striking a door several times in a clear repeated pattern.
Correct trigger: knock
Why: The repeated pattern makes it obvious that someone is knocking.
Incorrect trigger chosen: hit
Why this is wrong: Hit would only describe one strike, not the repeated knocking you hear.
Example 2: You hear a loud crash followed by pieces scattering across the floor.
Correct trigger: break
Why: The crash and the scattered sounds together make it clear that something has broken.
Incorrect trigger chosen: thud
Why this is wrong: Thud would describe one dull noise, not an object breaking.
Example 3: You hear someone beginning to whimper and then crying steadily with sobs between breaths.
Correct trigger: cry
Why: The sound is clearly someone crying.
Incorrect trigger chosen: breathe
Why this is wrong: Breathing is only a background sound and not the actual action you hear.

\end{tcolorbox}
\caption{Annotator instructions for audio narrative-level event trigger detection.}
\label{fig:audio-etd-instructions}
\end{figure*}

\begin{figure*}[t]
\centering
\begin{tcolorbox}[colback=blackcolor,colframe=black!75!white, title=Dialogue Narrative-Level Event Trigger Detection]
 
A dialogue narrative event trigger is a single verb that names the real-world action or situation the speaker is talking about right now. It is the actual thing that happened (or is happening) in the story, not the act of speaking itself. The trigger should be the key event the line reveals, admits, accuses, or makes real. Test it by replacing the whole line with trigger and if the story still feels the same, it's correct.

Example 1: A man says, ``I know I messed up. I am sorry for hurting you.''  
Correct trigger: hurt  
Why: He is owning up to hurting the listener.  
Incorrect trigger: apologize  
Why it is wrong: ``Apologize'' is what he's doing with words; ``hurt'' is what he actually did.

Example 2: A woman says, ``Please tell me where you went last night.''  
Correct trigger: go  
Why: The whole point of her question is the hidden trip last night; ``go'' is the real event she's trying to uncover.  
Incorrect trigger: ask  
Why it is wrong: ``Ask'' is the speech action; ``go'' is the thing she cares about.

Example 3: A teenager says, ``You never listen to me and you broke your promise again.''  
Correct trigger: break 
Why: The teenager is pointing at the broken promise as the real problem.  
Incorrect trigger: accuse  
Why it is wrong: ``Accuse'' is how it's being said; ``break promise'' is what was done.

Example 4: A woman says, ``I cannot stay here anymore. I am leaving you.''  
Correct trigger: leave  
Why: With those words she is making the leaving happen right now.  
Incorrect trigger: stay  
Why it is wrong: ``Stay'' is the opposite of what she's doing. 
 
\end{tcolorbox}
\caption{Annotator instructions for dialogue narrative-level event trigger detection.}
\label{fig:dialogue-etd-instructions}
\end{figure*}

\begin{figure*}[t]
\centering
\begin{tcolorbox}[colback=blackcolor,colframe=black!75!white, title=Narrative-Level Event Argument Extraction - Part 1]
 
After you select the event trigger, the platform shows you a list of PropBank arguments for that verb. These arguments represent the possible participants or elements of the event, such as who performs the action, who is affected, what object is involved or what causes the event. You do not invent new arguments or rewrite anything. You only choose from the arguments already shown in the list. Your task is to select only the arguments that correctly describe what is happening in the video, audio or dialogue. Every argument is optional but strongly encouraged to be filled. You fill the argument with information that is clearly present in at least one modality. If the trigger is kill, you would select the killer and the victim if you can identify them. If the trigger is arrest, you would select the officer and the suspect. If the trigger is warn, you would select the person giving the warning and the person receiving it. The arguments must always match what is actually shown or said.
Arguments can come from any source. A visual event can use information from dialogue or audio if those elements help identify who did what or explain what caused the action. This is allowed as long as the information exists in the story. Arguments can also come from another scene. For example, if the kill event happens early in the movie but the killer's identity is revealed much later, you may fill the killer argument using the later scene as long as you indicate the correct scene ID. When filling an argument from a different scene, you must add the scene ID in respective field.
Once you choose the arguments, read the event as a simple sentence using the trigger and the selected roles. If this description accurately matches what happened, the annotation is correct. If something does not fit, the problem usually lies in the trigger choice. In that case you return to the trigger selection, choose a better verb and then fill the arguments again.
 
\end{tcolorbox}
\caption{Annotator instructions for narrative-level event argument extraction (Part 1).}
\label{fig:eae-instructions-part1}
\end{figure*}

\begin{figure*}[t]
\centering
\begin{tcolorbox}[colback=blackcolor,colframe=black!75!white, title=Narrative-Level Event Argument Extraction - Part 2]

Example 1 (Visual Event with Visual and Dialogue Support)
A man grabs another person from behind and strangles them until they collapse.
The chosen verb is kill.
The system provides arguments such as killer, victim and cause.
Filled arguments:
	killer: Adam (the man in a dark jacket who performs the strangling) (from the visual scene)
	victim: John (the person in a grey shirt who collapses) (from the visual scene)

Argument source note:
Even if John's name is spoken in dialogue in another scene, you may still use that information to fill the argument. Cross-scene filling is allowed if the information is explicitly revealed later.
Example 2 (Visual Event Supported by Audio and Dialogue)
Two officers chase a suspect behind a building. In the visual scene we see the officers catch him. In the audio we hear handcuffs clicking. In dialogue one officer says, ``We finally got him.''

The chosen verb is arrest.
Filled arguments:
officer: Officer Miller and Officer Perez (two uniformed officers who catch the suspect) (from the visual scene)
	suspect: Daniel (the man in a black hoodie being restrained) (from dialogue and visual scene)
	place: behind the old brick warehouse (a red brick two story building with metal fire stairs) (from the visual scene)
Argument source note:
 Dialogue confirms the identity of the suspect even though the arrest action is visual.
Example 3 (Dialogue Event with Visual and Audio Support)
A woman faces her friend. She raises her hand in concern. The audio reveals loud creaks from a nearby bridge. She says, ``Stay back. It could collapse.''
The chosen verb is warn.

Filled arguments:
warner: Lisa (the woman raising her hand and giving the warning) (from the visual scene and dialogue)
	warnee: Emily (the friend standing in front of her) (from the visual scene)
	cause: the unstable wooden bridge (wooden planks bending and creaking over the river) (from audio and visual)
Argument source note:
The event draws information from all three modalities. The spoken line conveys the intention, audio reveals the danger and visuals identify who is warning whom.

Example 4 (Audio Event with Later Visual ID)
A loud scream echoes from another room. No one is shown, but later in scene 12 the movie reveals who screamed.
 The chosen verb is scream.
Filled arguments:
	screamer: Chloe (identified later when she says ``I was the one who screamed earlier'') (from scene 12, dialogue)
	place: the upstairs hallway (from the audio echo description and later visual scene)

Argument source note:
 Even though the scream is heard in scene 3, the identity is filled using information revealed later in scene 12.

Example 5 (Dialogue Event Identified Through Speech Alone)
A man says, ``I promise I will fix everything.''
The chosen verb is promise.
Filled arguments:
	promiser: Mark (the man speaking) (from the dialogue)
	promisee: Claire (standing next to him and receiving the promise) (from the visual scene)
Argument source note:
 The event itself comes from the spoken line, but the visual scene identifies the listener.
 
\end{tcolorbox}
\caption{Annotator examples for narrative-level event argument extraction (Part 2).}
\label{fig:eae-instructions-part2}
\end{figure*}

\begin{figure*}[t]
\centering
\begin{tcolorbox}[colback=blackcolor,colframe=black!75!white, title=Narrative-Level Event Relation Extraction]
 
A relation annotation identifies how two events are connected within the story based only on the provided plot, context and argument information. You do not create new relation types. You select from the relation list that is shown to you. A valid relation must be supported by explicit narrative details. Examples include temporal order, causal impact, preconditioned setup and hierarchical containment. Every relation must reflect a real link between the two events rather than an imagined one.
Once the system shows you the two events and the allowed relation labels, review each event in its context. Look at the verbs, the arguments and the narrative descriptions. Select a relation only when the plot clearly shows a connection. If an event directly triggers another, you choose a causal relation. If one event creates conditions that make the other possible without forcing it, you choose a preconditioned relation. If the events overlap and one is part of the other, you choose a hierarchical relation. If they only follow each other in time with no deeper link, you choose a temporal relation. If no link is supported, you select no relation.
After selecting a relation, confirm that it fits the way the two events work together in the story. You should be able to point to specific details that justify the relation. If the two events cannot be read as connected through the chosen label, you must revise the decision or select no relation. The correct choice is always the weakest option that fits the evidence. If any uncertainty remains and you cannot support the connection with clear details, the correct answer is no relation.

Example 1: Causal Relation (Visual Event + Dialogue)
Event 1: kill
 killer: Adam
 victim: John
Event 2: report
 speaker: a witness
 content: ``He killed him''
Correct relation: causal
 Why: The killing is the reason the witness reports it. The dialogue event arises because of the visual event.
Incorrect relation chosen: temporal
 Why incorrect: The report does not merely follow the killing. It exists because of it.

Example 2: Preconditioned Relation (Dialogue + Visual)
Event 1: ask
 speaker: Emily
 addressee: Frank
 content: ``Please open the door''
Event 2: open
 agent: Frank
 theme: wooden door
Correct relation: preconditioned
 Why: Emily's request allows or invites the opening but does not force it. Frank still chooses to act.
Incorrect relation chosen: causal
 Why incorrect: Requests do not automatically cause compliance.

\end{tcolorbox}
\caption{Annotator instructions and examples for narrative-level event relation extraction.}
\label{fig:ere-instructions}
\end{figure*}

\definecolor{jsonback}{rgb}{0.95,0.95,0.95}

\lstdefinestyle{jsonbox}{
    basicstyle=\ttfamily\small,
    breaklines=true,
    breakatwhitespace=true,
    columns=flexible,
    keepspaces=true,
    showstringspaces=false,
    frame=none,
}

\begin{figure*}[!t]
\centering

\begin{tcolorbox}[colback=jsonback, colframe=black!75!white, 
    title={Task 1: Visual Narrative Event Trigger Detection}, 
    fonttitle=\bfseries,
    width=\textwidth,
    left=5pt, right=5pt, top=5pt, bottom=5pt]
\begin{lstlisting}[style=jsonbox]
{
  "messages": [
    {
      "role": "user",
      "content": "Visual Narrative Event Trigger Detection: Narrative event triggers describe the real story event taking place in the scene, not the small physical motions happening on the surface. A scene may contain many tiny actions such as walking, sitting, turning, picking something up or opening a door. These movements show how someone moves, but they do not explain what the moment means. A narrative event is the larger action that these motions add up to. Someone standing up, taking a suitcase and walking out is not performing three events and together these actions mean the person leaves. Someone stepping forward, pointing and blocking another person's path adds up to confront. Someone gathering belongings and closing a door behind them adds up to leave. A single narrative event can contain several visible motions, yet the motions themselves are not the event. To find the correct narrative event, first observe the physical actions without interpreting them. Then consider what these motions together accomplish in the story. Once the story-level action is clear, select the verb that best expresses this meaning. You must always choose a concrete visual narrative event rather than an abstract internal state such as think, realize or feel. A concrete event is something that a viewer can see happening directly in the video. Now, identify the narrative event trigger in this video."
    },
    {
      "role": "assistant",
      "content": "TRIGGER"
    }
  ],
  "videos": ["output_fps1.mp4"]
}
\end{lstlisting}
\end{tcolorbox}
\caption{Prompt template for visual narrative event trigger detection.}
\label{fig:etd-prompt-template}
\end{figure*}

\begin{figure*}[!t]
\centering

\begin{tcolorbox}[colback=jsonback, colframe=black!75!white, 
    title={Task 2: Video Event Argument Extraction}, 
    fonttitle=\bfseries,
    width=\textwidth,
    left=5pt, right=5pt, top=1pt, bottom=1pt]
\begin{lstlisting}[style=jsonbox]
{
  "messages": [
    {
      "role": "user",
      "content": "Video Event Argument Extraction: You will see a list of arguments for the given event trigger. These arguments represent the possible participants or elements of the event, such as who performs the action, who is affected, what object is involved or what causes the event. You do not invent new arguments or rewrite anything. You only choose from the arguments already shown in the list. Your task is to select only the arguments that correctly describe what is happening in the video, audio or dialogue. Every argument is optional but strongly encouraged to be filled. You fill the argument with information that is clearly present in at least one modality. Arguments can come from any source. A visual event can use information from dialogue or audio if those elements help identify who did what or explain what caused the action. Arguments can also come from another scene. When filling an argument from a different scene, you must add the scene ID in respective field. Once you choose the arguments, read the event as a simple sentence using the trigger and the selected roles. If this description accurately matches what happened, the annotation is correct. Now, extract the event arguments for the trigger 'kill' in this video."
    },
    {
      "role": "assistant",
      "content": "ARG_ROLE_1: Visual Description 1 (modality source), ARG_ROLE_2: Visual Description 2 (modality source)"
    }
  ],
  "videos": ["output_fps1.mp4"]
}
\end{lstlisting}
\end{tcolorbox}
\caption{Prompt template for video event argument extraction.}
\label{fig:eae-prompt-template}
\end{figure*}

\begin{figure*}[!t]
\centering

\begin{tcolorbox}[colback=jsonback, colframe=black!75!white, 
    title={Task 3: Video Relation Extraction}, 
    fonttitle=\bfseries,
    width=\textwidth,
    left=5pt, right=5pt, top=1pt, bottom=1pt]
\begin{lstlisting}[style=jsonbox]
{
  "messages": [
    {
      "role": "user",
      "content": "Video Relation Extraction: A relation annotation identifies how two events are connected within the story based on the context and argument information. You do not create new relation types. You select from the relation list that is shown to you. A valid relation must be supported by explicit narrative details. Examples include temporal order, causal impact, preconditioned setup and hierarchical containment. Every relation must reflect a real link between the two events rather than an imagined one. Review each event in its context. Look at the verbs, the arguments and the narrative descriptions. Select a relation only when there is a clear connection. If an event directly triggers another, you choose a causal relation. If one event creates conditions that make the other possible without forcing it, you choose a preconditioned relation. If the events overlap and one is part of the other, you choose a hierarchical relation. If they only follow each other in time with no deeper link, you choose a temporal relation. If no link is supported, you select no relation. After selecting a relation, confirm that it fits the way the two events work together in the story. The correct choice is always the weakest option that fits the evidence. If any uncertainty remains and you cannot support the connection with clear details, the correct answer is no relation. Now, identify the relation between Event 1 and Event 2 in this video."
    },
    {
      "role": "assistant",
      "content": "RELATION_TYPE"
    }
  ],
  "videos": ["output_fps1.mp4"]
}
\end{lstlisting}
\end{tcolorbox}
\caption{Prompt template for video event relation extraction.}
\label{fig:ere-prompt-template}
\end{figure*}

\begin{figure*}[!t]
\centering
\vspace{0.5em}

\begin{tcolorbox}[colback=jsonback, colframe=black!75!white, 
    title={Task 4: Video Event Localization}, 
    fonttitle=\bfseries,
    width=\textwidth,
    left=5pt, right=5pt, top=5pt, bottom=5pt]
\begin{lstlisting}[style=jsonbox]
{
  "messages": [
    {
      "role": "user",
      "content": "Video Event Localization: Video event localization identifies the temporal boundaries of narrative events within the video. You must determine when each narrative event begins and ends based on the visual, audio and dialogue cues present in the video. Identify the narrative event trigger in the video. Observe the small physical actions that compose the narrative event. Mark the start time when the first action contributing to the narrative event begins. Mark the end time when the last action contributing to the narrative event concludes. Verify that the temporal boundaries capture the complete narrative event. The temporal boundaries must encompass all small actions that combine into the narrative event. Start and end times should be precise and based on observable visual or audio cues. Do not include actions that occur before or after the narrative event. The localization must align with the narrative event identified, not individual motions. Use timestamps in the format [start_time, end_time] in seconds. Now, identify the temporal boundaries of the narrative event in this video."
    },
    {
      "role": "assistant",
      "content": "[START_TIME, END_TIME]"
    }
  ],
  "videos": ["output_fps1.mp4"]
}
\end{lstlisting}
\end{tcolorbox}
\caption{Prompt template for video event localization.}
\label{fig:el-prompt-template}
\end{figure*}

\begin{figure*}[t]
\centering
\begin{tcolorbox}[colback=blackcolor,colframe=black!75!white, title=Judge Template]

\{DEFINITION\}

\{RULES\}

\{OUTPUT FORMAT RULE\}

\# Ground Truth: \{GT INPUTS\}

\# Prediction: \{MODEL PREDICTIONS\}

\end{tcolorbox}
\caption{Generic judge template used for LLM-based evaluation.}
\label{fig:judge-template}
\end{figure*}

\begin{figure*}[t]
\centering
\begin{tcolorbox}[colback=blackcolor,colframe=black!75!white, title=ETD Judge template]

DEFINITION: Visual Narrative Event Trigger Detection: Narrative event triggers describe the real story event taking place in the scene, not the small physical motions happening on the surface. A scene may contain many tiny actions such as walking, sitting, turning, picking something up or opening a door. These movements show how someone moves, but they do not explain what the moment means. A narrative event is the larger action that these motions add up to. Someone standing up, taking a suitcase and walking out is not performing three events and together these actions mean the person leaves. Someone stepping forward, pointing and blocking another person's path adds up to confront. Someone gathering belongings and closing a door behind them adds up to leave. A single narrative event can contain several visible motions, yet the motions themselves are not the event. To find the correct narrative event, first observe the physical actions without interpreting them. Then consider what these motions together accomplish in the story. Once the story-level action is clear, select the verb that best expresses this meaning. You must always choose a concrete visual narrative event rather than an abstract internal state such as think, realize or feel. A concrete event is something that a viewer can see happening directly in the video. Now, identify the narrative event trigger in this video. Return your answer as a JSON object with keys: "verb" and "context".
\newline

RULES: 
\# Evaluation:
Decide if prediction describes the same narrative event as the ground truth. Synonyms and paraphrases are allowed. If the meaning/synonym of the trigger is preserved, it is correct. 
\newline

OUTPUT FORMAT RULE: \# Output format: Follow the output format strictly: return a json object with keys "verdict" and "scores". The "verdict" key will contain one of the two values: 1 if correct, 0 if incorrect. This is your final evaluation verdict. The "scores" key will contain a value between 0 and 1, which is your confidence that the "verdict" is 1.
 
\end{tcolorbox}
\caption{LLM judge prompt template for event trigger detection.}
\label{fig:etd-judge-template}
\end{figure*}

\begin{figure*}[t]
\centering
\begin{tcolorbox}[colback=blackcolor,colframe=black!75!white, title=EAE Judge template]

DEFINITION: Video Event Argument Extraction: Fill in the values for the following semantic roles based on what you observe in the video. Return your answer as a JSON object with the role names as keys and the observed values from the video.
\newline

RULES: 
\# Evaluation:
Decide if predicted arguments match in meaning as the ground truth. Minor name differences are allowed.
\newline

OUTPUT FORMAT RULE: \# Output format: Follow the output format strictly: return a json object with keys "verdict" and "scores". The "verdict" key will contain one of the two values: 1 if correct, 0 if incorrect. This is your final evaluation verdict. The "scores" key will contain a value between 0 and 1, which is your confidence that the "verdict" is 1.

\end{tcolorbox}
\caption{LLM judge prompt template for event argument extraction.}
\label{fig:eae-judge-template}
\end{figure*}
\end{document}